\documentclass[manuscript, screen, sigconf]{acmart}

\usepackage{colortbl}
\usepackage{dsfont}

\usepackage{tabularx}
\usepackage{multirow}

\definecolor{oursrow}{HTML}{E6F2F4}

\definecolor{green}{HTML}{008015}

\newcommand{\etal}{\textit{et al. }}

\AtBeginDocument{%
  }

\copyrightyear{2023}
\acmYear{2023}
\setcopyright{rightsretained}
\acmConference[ICMR '23]{International Conference on Multimedia Retrieval}{June 12--15, 2023}{Thessaloniki, Greece}
\acmBooktitle{International Conference on Multimedia Retrieval (ICMR '23), June 12--15, 2023, Thessaloniki, Greece}\acmDOI{10.1145/3591106.3592262}
\acmISBN{979-8-4007-0178-8/23/06}

\acmSubmissionID{5151}

\begin{document}

\title{Not Only Generative Art: Stable Diffusion\\for Content-Style Disentanglement in Art Analysis}

\author{Yankun Wu}
\affiliation{%
  \institution{Osaka University}
  \city{}
  \country{}
  }
\email{yankun@is.ids.osaka-u.ac.jp}

\author{Yuta Nakashima}
\affiliation{%
  \institution{Osaka University}
  \city{}
  \country{}
  }
\email{n-yuta@ids.osaka-u.ac.jp}

\author{Noa Garcia}
\affiliation{%
  \institution{Osaka University}
  \city{}
  \country{}
  }
\email{noagarcia@ids.osaka-u.ac.jp}

\renewcommand{\shortauthors}{Wu et al.}

\begin{abstract}
The duality of content and style is inherent to the nature of art. For humans, these two elements are clearly different: content refers to the objects and concepts in the piece of art, and style to the way it is expressed. This duality poses an important challenge for computer vision. The visual appearance of objects and concepts is modulated by the style that may reflect the author's emotions, social trends, artistic movement, etc., and their deep comprehension undoubtfully requires to handle both. A promising step towards a general paradigm for art analysis is to disentangle content and style, whereas relying on human annotations to cull a single aspect of artworks has limitations in learning semantic concepts and the visual appearance of paintings. We thus present GOYA, a method that distills the artistic knowledge captured in a recent generative model to disentangle content and style. Experiments show that synthetically generated images sufficiently serve as a proxy of the real distribution of artworks, allowing GOYA to separately represent the two elements of art while keeping more information than existing methods.
\end{abstract}




\maketitle

%%%%%%%%% BODY TEXT

% \input{01-intro.tex}
% \input{02-relatedwork.tex}
% \input{03-preliminaries.tex}
% \input{04-method.tex}
% \input{05-experiments.tex}
% \input{06-conclusion.tex}

\section{Introduction}

Content and style are two fundamental elements in the analysis of art. On the one hand, \textit{content} describes the concepts depicted in the image, such as the objects, people, or locations. It addresses the question \textit{what the artwork is about}. On the other hand, style describes the visual appearance of the image: its color, composition, or shape, addressing the question \textit{how the artwork looks}. Through a unique combination of content and style, a piece of art looks as it is, making the disentanglement of these two elements an essential trait in the study of digital humanities. 

\begin{figure}
% \vspace{10pt}
\hspace{-25pt}
    \centering
    \includegraphics[width=1\columnwidth]{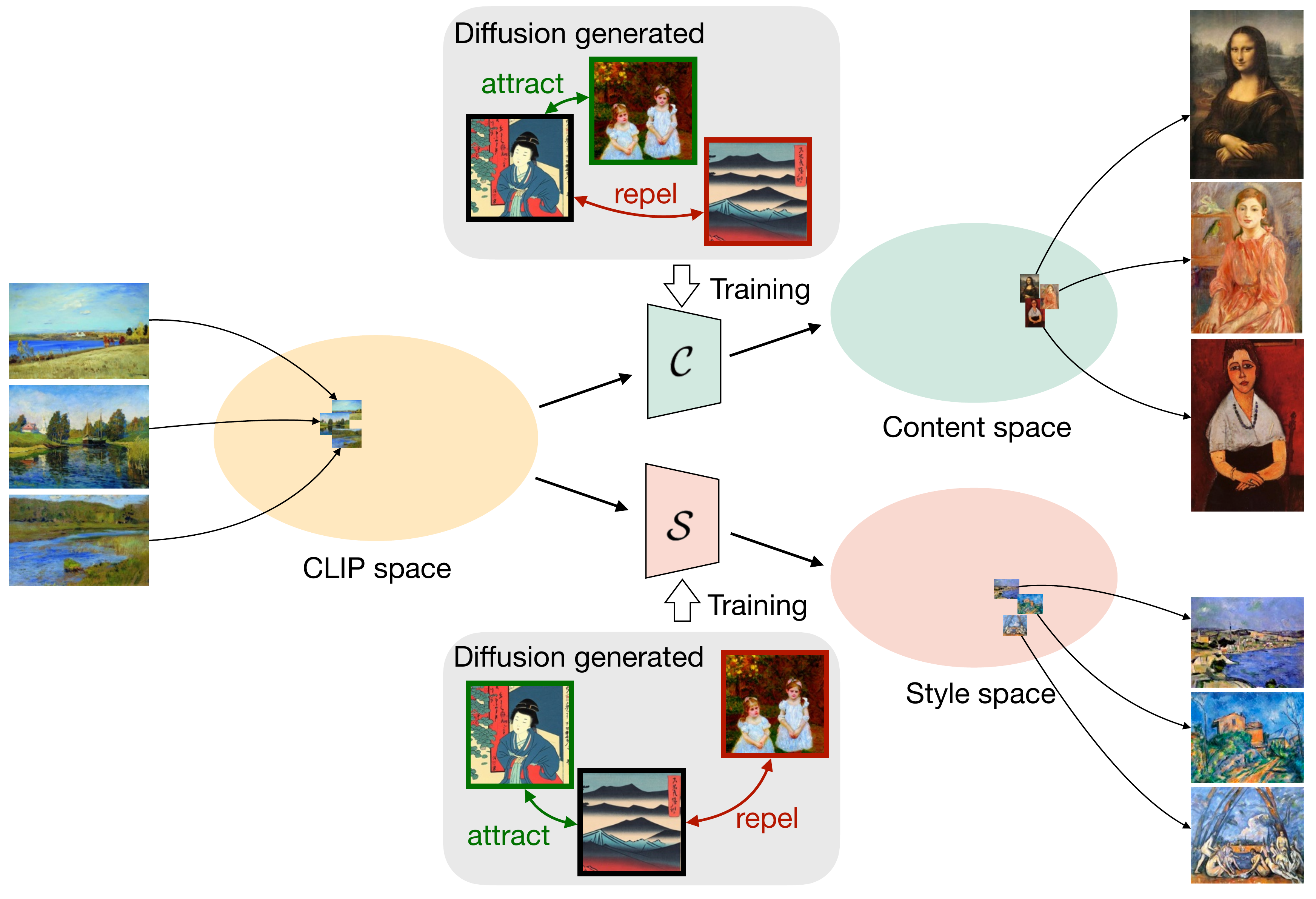}
    \vspace{-15pt}
    \caption{An overview of our method GOYA. By using Stable Diffusion generated images, we disentangle content and style spaces from CLIP space, where content space represents semantic concepts and style space captures visual appearance. }
    \label{fig:intro}
    \vspace{-3pt}
\end{figure}

Whereas for humans, content and style are easily distinguishable (we can often tell apart the topics depicted in a painting from their visual appearance without much trouble), the boundary is not so clear from a computer vision perspective. Traditionally, for analyzing content in artworks, computer vision relies on object recognition techniques \cite{carneiro2012artistic,crowley2014state,gonthier2018weakly}. However, even when artworks contain similar objects, the  \textit{subject matter} may still be different. Likewise, the automatic analysis of style is not without controversy. As there is not a formal definition of what visual appearance is, there is a degree of vagueness and subjectivity in the computation of style. Some methods \cite{garcia2019context, cetinic2018fine} rely on well-established attributes, such as author or artistic movement, as a proxy to classify style. While this definition may be enough for some applications, such as artist identification \cite{van2015toward}, it is not appropriate in others, e.g. style transfer \cite{gatys2016image} or image search \cite{ypsilantis2021met}. In style transfer, for example, style is defined as the low-level features of an image (e.g. colors, brushstrokes, shapes). In a broader sense, however, style is not formed by a single image but by a set of artworks that share a common visual appearance \cite{lang2018reflecting}.

On top of these challenges, most of the methods for art analysis are trained with full supervision, requiring each image in the dataset to be annotated with its corresponding content or style category. This categorization poses some additional problems. Firstly, although there are digitized collections of artworks that can be leveraged for supervised learning (e.g. WikiArt\footnote{\url{https://www.wikiart.org/}}, The Met\footnote{\url{https://www.metmuseum.org/}}, Web Gallery of Art\footnote{\url{https://www.wga.hu/}}), labels for new artworks are not straightforward to obtain, and often require experts to annotate them. Secondly, the annotated labels, which are often single-words, reflect only some general traits of a set of artworks while ignoring the subtle properties of each image. For instance, given a painting with a genre label \textit{still life} and an artistic movement label \textit{Expressionism}, what scene does it depict and how does its visual appearance look like? We can infer some of the coarse attributes it may carry, e.g. inanimate subjects from \textit{still life} and strong subjective emotions from \textit{Expressionism}. However, some fine attributes such as depicted concepts, color composition, and brushstrokes still remain obscure. When training based on labels, it is difficult to learn the subtle content and style discrepancy in images.
To overcome the limitations imposed by labels, some work \cite{garcia2018read,cetinic2021towards,bai2021explain} has relied on natural language descriptions instead of categorical classes. Although natural language can contribute to resolving the ambiguity and rigidness of labels, human experts still need to write descriptions for each artwork. 

Moving away from human supervision, we investigate the generative power of a popular text-to-image model, Stable Diffusion \cite{rombach2022high}, and propose to leverage the distilled knowledge as a prior to learn disentangled content and style embeddings of paintings. 
Given a text input, also known as prompt, Stable Diffusion can generate a set of diverse synthetic images while maintaining content and style consistency. The subtle characteristics of content and style in the synthetically generated images can be controlled by well-defined prompts. Thus, we propose to use the generated images with the help of the input prompts to train a model that disentangles content and style in paintings without direct human annotations. 
Concurrent to our work, it has been shown that using Stable Diffusion generated images can be useful for image classification tasks \cite{sariyildiz2023fake}.

The intuition behind our method, named GOYA (disentan\underline{G}lement of c\underline{O}ntent and st\underline{Y}le with gener\underline{A}tions), is that although there is no explicit boundary between different contents or styles, we can distinguish significant dissimilarities by comparison. Our simple yet effective model (Figure \ref{fig:intro}) first extracts joint content-style embeddings using pre-trained CLIP image encoder \cite{radford2021learning}, and then applies two independent content and style transformation networks to learn disentangled content and style embeddings. As mentioned before, the weights of the transformation networks are trained on the generated synthetic images with contrastive learning, reducing the need of using human image-level annotations. 

We conduct three tasks and an ablation study on a popular benchmark of paintings, the WikiArt dataset \cite{artgan2018}. We show that GOYA, by distilling the knowledge from Stable Diffusion generated images, disentangles content and style better than models trained on real paintings. Moreover, experiments show that the resulting disentangled spaces are useful for downstream tasks such as similarity retrieval and art classification. 
In summary, our contributions are:
\begin{itemize}
    \item We design a model to disentangle content and style information from pre-trained CLIP's latent space.
    \item We propose to train the disentanglement model with synthetically generated images, instead of real paintings, via Stable Diffusion and prompt design.
    \item We show that the information in Stable Diffusion generated images can be effectively distilled for art analysis, performing well on tasks such as art retrieval and art classification.
\end{itemize}
\vspace{-5pt}
Our findings open the way for adopting generative models in digital humanities, not only for generation but also for analysis.

\begin{figure*}
\hspace{-25pt}
    \centering
    \includegraphics[width=0.9\textwidth]{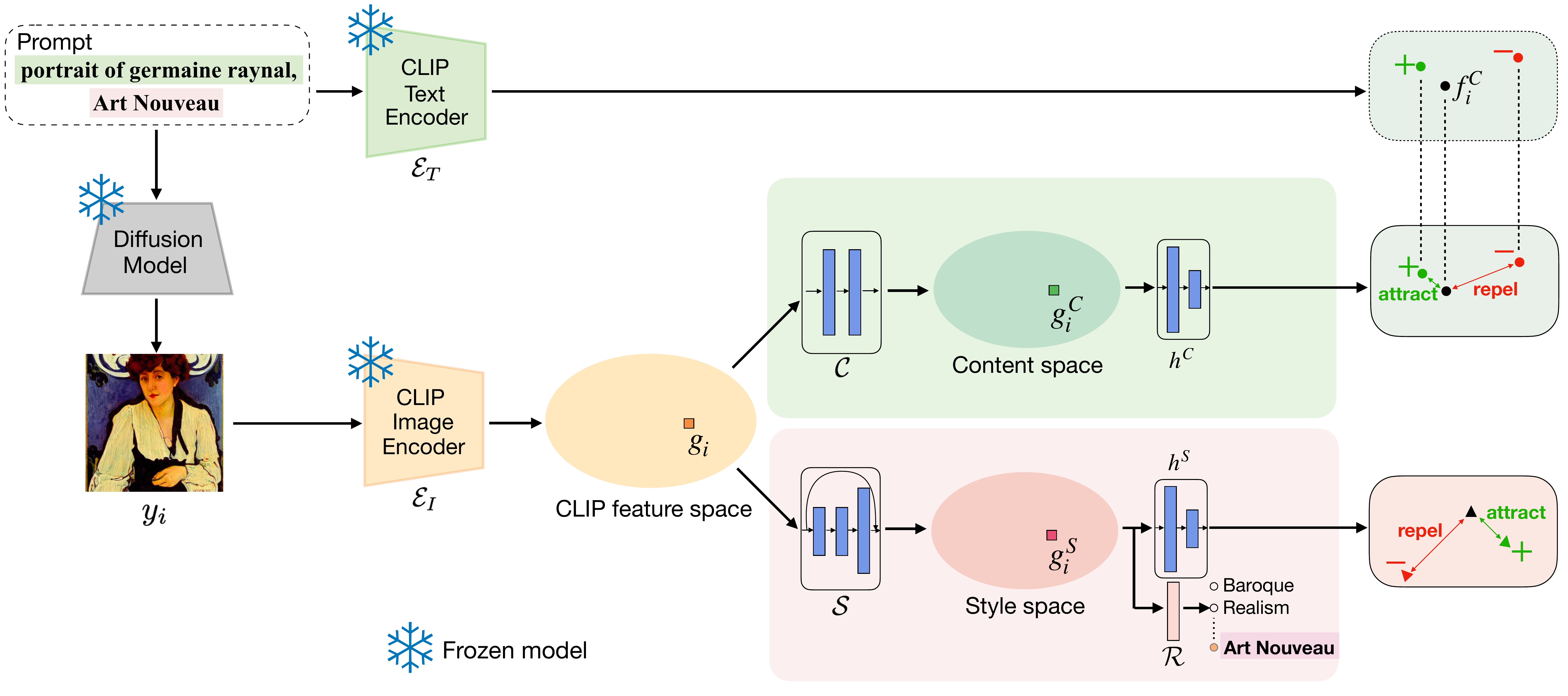}
    \caption{Details of our proposed method GOYA for content and style disentanglement. Given a synthetic prompt containing content (first part of the prompt, in green) and style (second part of the prompt, in red) descriptions, we generate synthetic diffusion images. We compute CLIP embeddings with the frozen CLIP image encoder, and generate content and style disentangled embeddings with two dedicated encoders $\mathcal{C}$ and $\mathcal{S}$, respectively. In the training stage, projectors $h^C$ and $h^S$ and style classifier $\mathcal{R}$ are used to train GOYA with contrastive learning. For content, contrastive learning pairs are chosen based on the text embedding of content description in the prompt extracted by frozen CLIP text encoder. For style, contrastive learning pairs are chosen based on the style description in the prompt.}
    \label{fig:GOYA}
    \vspace{-2pt}
\end{figure*}

\section{Related work}
\paragraph{Art analysis}
The use of computer vision techniques for art analysis has been an active research topic for decades, in particular in tasks such as attribute classification \cite{mensink2014rijksmuseum, tan2016ceci, ma2017part, garcia2019context, el2021gcnboost}, object recognition \cite{carneiro2012artistic,crowley2014state,gonthier2018weakly,shen2019discovering}, or image retrieval \cite{carneiro2012artistic, crowley2014state, garcia2018read, ypsilantis2021met}. Fully-supervised tasks (e.g., genre or artist classification \cite{tan2016ceci, mao2017deepart, garcia2020contextnet}) have obtained outstanding results by leveraging neural networks trained on annotated datasets \cite{mensink2014rijksmuseum,saleh2016large,mao2017deepart,strezoski2018omniart}. However, image annotations have some limitations.
An important limitation is the categorization of styles. Multiple datasets \cite{karayev2013recognizing,mensink2014rijksmuseum,florea2016pandora,saleh2016large,mao2017deepart,wilber2017bam,strezoski2018omniart,khan2021stylistic} provide style labels, which have been leveraged by abundant work \cite{van2015toward,lecoutre2017recognizing,chu2018image,elgammal2018shape,sabatelli2018deep,chen2019recognizing,sandoval2019two} for style classification. This direction of work assumes style to be a static attribute, instead of dynamic and evolving \cite{lang2018reflecting}. 
A different interpretation is provided in style transfer \cite{gatys2016image, deng2020arbitrary, chen2021diverse, chen2021dualast}. A model extracts the low-level representation of a \textit{stylized image} (e.g. a painting) and applies it to a \textit{content image} (e.g. a plain photograph), defining style by a single artwork, i.e. the color, shape, and brushstroke of the stylized image. To overcome the rigidness of labels in supervised learning and the narrowness of a single image in style transfer, we propose learning disentangled embeddings of content and style by similarity comparisons by leveraging the flexibility of a text-to-image generative model.

\paragraph{Representation disentanglement}
Learning disentangled representation plays an essential role in various computer vision tasks such as style transfer \cite{kotovenko2019content, xie2022artistic}, image manipulation \cite{shi2022semanticstylegan, xu2022predict}, and image-to-image translation \cite{yu2019multi, gonzalez2018image, gabbay2020improving, huang2018multimodal}. The goal is to discover discrete factors of variation in data, thus improving the interpretability of representations and enabling a wide range of downstream applications. To disentangle attributes like azimuth, age, or gender, previous work have built on adversarial learning \cite{chen2016infogan, denton2017unsupervised, tran2017disentangled, khrulkov2021disentangled} or variational autoencoders (VAE) \cite{kingma2013auto, higgins2016beta, shao2020controlvae}, aiming to encourage discrete properties in a single latent space. 
For content and style disentanglement, several work apply generative model \cite{kotovenko2019content, kazemi2019style, gabbay2020improving}, diffusion model \cite{kwon2022diffusion, wu2022uncovering}, or an autoencoder architecture with contrastive learning \cite{kotovenko2019content, collomosse2017sketching, ruta2021aladin}. In the art domain, ALADIN \cite{ruta2021aladin} concatenates the adaptive instance normalization (AdaIN) \cite{huang2017adain} feature into the style encoder to learn style embedding for visual searching. Kotovenko \etal\cite{kotovenko2019content} propose fixpoint triplet loss and disentanglement loss for performing better style transfer. However, these work lack semantic analysis of content embeddings in paintings. Recently, Vision Transformer (ViT) \cite{dosovitskiy2020image}-based models show the ability to obtain structure and appearance embeddings \cite{kwon2022diffusion, tumanyan2022splicing, caron2021emerging}. DiffuseIT \cite{kwon2022diffusion} and Splice \cite{tumanyan2022splicing} learn content and style embeddings by utilizing the keys and the global $[\mathrm{CLS}]$ token of pre-trained DINO \cite{caron2021emerging}. In our work, taking advantage of the generative model, our approach builds a simple framework to decompose the latent space into content and style spaces with contrastive learning, exploring to employ generated images on representation learning.

\paragraph{Text-to-image generation}
Text-to-image models intend to perform synthetic image generation conditioned on a given text input. Catalyzed by datasets with massive text-image pairs that have emerged in recent years, many powerful text-to-image generation models sprung up \cite{rombach2022high, li2022stylet2i, liao2022text, ding2021cogview, ramesh2022hierarchical, tao2022df, saharia2022photorealistic}. For instance, CogView \cite{ding2021cogview} is trained on 30 million text-image pairs while DALL-E 2 \cite{ramesh2022hierarchical} is trained on 650 million text-image pairs. 
One of their main challenges is achieving semantic coherence between guiding texts and generated images. This has been addressed by using pre-trained CLIP embeddings \cite{radford2021learning} to construct aligned text and image features in the latent space \cite{zhou2022towards, li2022stylet2i, kwon2022clipstyler}. 
Another challenge is to obtain high-resolution synthetic images. GAN-based models \cite{liao2022text, tao2022df, zhou2022towards, tan2020kt} have achieved good performance in improving the quality of generated images, however, they suffer from instability in training. Exploiting the superiority of training stability, work based on Diffusion models \cite{rombach2022high, saharia2022photorealistic, ramesh2022hierarchical} have recently become a popular tool for generating near-human quality images. Despite the rapid development of models for image generation, how to leverage the feature of synthetic images remains an underexplored area of research. In this paper, we study the potential of generated images for enhancing representation learning.

\section{Preliminaries} \label{sec:preliminaries}

\subsection{Stable Diffusion} \label{sec:StableDiffusion}
Diffusion models \cite{ho2020denoising, rombach2022high} are generative methods trained on two stages: a forward process with a Markov chain to transform input data to noise, and a reversed process to reconstruct data from the noise, obtaining high-quality performance on image generation. 

To reduce the training cost and accelerate the inference process, Stable Diffusion \cite{rombach2022high} trains the diffusion process in the latent space instead of the pixel space.
Given a text prompt as input condition, the text encoder transforms the prompt to a text embedding. Then, by feeding the embedding into the UNet through a cross-attention mechanism, the reversed diffusion process generates an image embedding in the latent space. Finally, the image embedding is fed to the decoder to generate a synthetic image.

In this work, we define symbols as follows: given a text prompt $x = \{x^C, x^S\}$ as input, we can obtain the generated image $y$. The text $x^C$ represents content description and $x^S$ denotes style description, where $\{\cdot\}$ indicates a comma-separated string concatenation.

\subsection{CLIP}
CLIP \cite{radford2021learning} is a text-image matching model that aligns text and image embeddings in the same latent space. It shows high consistency of the visual concepts in the image and the semantic concepts in the corresponding text. The text encoder $\mathcal{E}_T$ and image encoder $\mathcal{E}_I$ of CLIP are trained with 440 million text-image pairs, showing outstanding performance on various text and image downstream tasks, such as zero-shot prediction \cite{cheng2021data, zhang2022pointclip} and image manipulation \cite{li2022stylet2i, kwon2022clipstyler, kim2022diffusionclip}. Given the text $x$ and an image $y$, the CLIP embeddings $f$ from text, and $g$ from image, both in $\mathbb{R}^d$, can be computed as:
\begin{align}
  f &= \mathcal{E}_T(x),
  \label{eq:CLIP_text}\\
  g &= \mathcal{E}_I(y).
  \label{eq:CLIP_img}
\end{align}
% \vspace{-2pt}
To exploit the multi-modal CLIP space, we employ the pre-trained CLIP image encoder $\mathcal{E}_I$ to obtain CLIP image embeddings as the prerequisite for the subsequent disentanglement model. Moreover, during the training stage, the CLIP text embedding of a prompt is applied to acquire the semantic concepts of the generated image.

\section{GOYA}
The goal of our task is to learn disentangled content and style embeddings of artworks in two different spaces. Unlike previous work, we leverage the knowledge of generated images rather than real paintings, unlocking generative models for representation analysis.
Our idea is to borrow the Stable Diffusion's capability to generate a wide variety of images not only of diverse contents but also in various styles. Contrastive losses for content and style allow GOYA to learn the proximity of different artworks in the respective spaces with the consistency of generated images and text prompts.

% Figure is in intro.tex
Figure \ref{fig:GOYA} shows an overview of GOYA.
Given a mini-batch of $N$ prompts $\{x_i\}_{i=0}^{N}$, where $x_i=\{x^C_i, x^S_i\}$ with comma-connected content and style descriptions, we obtain diffusion generated images $y_i$ using Stable Diffusion. We then compute CLIP image embeddings $g_{i}$ by Eq.~(\ref{eq:CLIP_img}) and use a \textit{content encoder} and a \textit{style encoder} to obtain disentangled embeddings in two different spaces. As previous work has shown \cite{gatys2015neural} content and style have different properties, while content embeddings refer to higher layers in the deep neural network and style embeddings respond to lower layers. We design an asymmetric network architecture for extracting content and style, which is common in the art analysis domain \cite{mao2017deepart, kotovenko2019content, gatys2015neural, ruta2021aladin}.

\subsection{Content encoder}
The content encoder $\mathcal{C}$ maps CLIP image embedding $g_i$ to content embedding $g^C_i$ as:
\begin{equation}
    g_i^C = \mathcal{C}(g_i),
\end{equation}  
$\mathcal{C}$ is a two-layer perceptron (MLP) with ReLU non-linearity.
Following previous work \cite{chen2020simple}, to make content $g^{C}_{i}$ highly linear, at training time we add a non-linear projector $h^C$ on top of the content encoder, which is a three-layer MLP with ReLU non-linearity. 

\subsection{Style encoder}
Style encoder $\mathcal{S}$ also maps CLIP image embedding $g_i$ but to style embedding $g^S_i$ as:
\begin{equation}
    g_i^S = \mathcal{S}(g_i).
\end{equation}  
$\mathcal{S}$ is a three-layer MLP with ReLU non-linearity. In particular, following \cite{He2016DeepRL}, we apply a skip connection before the last ReLU non-linearity in $\mathcal{S}$.
Similar to the content encoder, non-linear projector $h^S$ with the same structure as $h^C$ is added after $\mathcal{S}$ to facilitate contrastive learning. 

\subsection{Content contrastive loss}
Unlike prior work \cite{kotovenko2019content}, which defines content similarity only when style-transferred images are from the same source, we use a broader definition of content similarity. We propose a soft-positive selection strategy that defines pairs of images with similar content according to their semantic similarity.
That is,  two images with similar semantic concepts are defined as a positive pair whereas images without semantic similarity are negative pairs.

To quantify \textit{semantic similarity} between a pair of images, we exploit the CLIP latent space and conduct text similarity between the associated texts. Given the content description $x^C_i$ of the image $y_{i}$, we assume that the CLIP text embedding $f^C_{i} = \mathcal{E}_T(x^C_{i})$ can be a proxy for the content of $y_i$. Therefore, given a pair of two diffusion images $(y_i, y_j)$ and a text similarity threshold $\epsilon^T$, they are a positive pair if $ D^T_{ij} \leq \epsilon^T$, where $D^T_{ij}$ is the text similarity obtained by the cosine distance between the CLIP text embedding $f^C_{i}$ and $f^C_{j}$. The content contrastive loss is defined as:
\begin{equation}
  L^{C}_{ij} = \mathds{1}_{[D^T_{ij} \leq \epsilon^T]}(1-D^C_{ij})  + \mathds{1}_{[D^T_{ij} > \epsilon^T]}\max(0, D^{C}_{ij}-\epsilon_c), 
\label{eq:content_loss}
\end{equation}
where $\mathds{1}_{[\cdot]}$ is the indicator function that gives 1 when the condition is true and 0 otherwise. $D^C_{ij}$ is the cosine distance between $h_C(g^C_i)$ and $h_C(g^C_j)$, which are the content embeddings of images after projection. $\epsilon_c$ is the margin that constrains the minimum distance of negative pairs. 

\subsection{Style contrastive loss}
The style contrastive loss is defined based on the style description $x^{S}$ given in the input prompt. If a pair of images shares the same style class, then they are considered to be a positive pair, which means that their style embeddings should be close in the style space. Otherwise, they are a negative pair, and they should be pushed away from each other. Given $(y_i, y_j)$, the style contrastive loss can be computed as:
\begin{equation}
  L^{S}_{ij} = \mathds{1}_{[x^{S}_{i}=x^{S}_{j}]}(1-D^S_{ij})) 
   + \mathds{1}_{[x^{S}_{i}\neq x^{S}_{j}]}\max(0, D^S_{ij}-\epsilon^S), 
\label{eq:style_contrastive_loss}
\end{equation}
where $D^S_{ij}$ is the cosine distance between the style embeddings $h^S(g^S_i)$ and $h^S(g^S_j)$ after projection, and $\epsilon^S$ is the margin.

\subsection{Style classification loss}
To learn the general attributes of each style, we introduce a style classifier $\mathcal{R}$ to predict the style description (given as $x^{S}_{i}$) based on the embedding $g^{S}_i$ of image $y_i$. Prediction $w^{S}_{i}$ by the classifier is given by:
\begin{equation}
    w^S_i = \mathcal{R}(g^S_i),
\end{equation}
where $\mathcal{R}$ is a linear layer network.
For training, we use softmax cross-entropy loss, which is denoted by $L^{SC}_i$.
Note that the training of this classifier does not rely on human annotations, but on the synthetic prompts and generated images by Stable Diffusion. 

\subsection{Total loss}
In the training process, we compute the sum of three losses. 
The overall loss function in a mini-batch is formulated as:
\begin{equation}
  L = \lambda^{C} \sum_{ij} L^{C}_{ij} + \lambda^S \sum_{ij} L^{S}_{ij} + \lambda^{SC}\sum_i L^{SC}_{i},
\label{eq:all_loss}
\end{equation}
where $\lambda^C$, $\lambda^S$ and $\lambda^{CS}$ are parameters to control the contributions of losses. We set $\lambda^C=\lambda^S=\lambda^{CS}=1$. The summantions over $i$ and $j$ are computed for all pairs of images in the mini-batch, and the summation over $i$ is for all images in the mini-batch.

\section{Evaluation}
We evaluate GOYA on three tasks: disentanglement (Section \ref{section:DC}), classification (Section \ref{section:classification}), and similarity retrieval (Section \ref{section:similarity}). We also conduct an ablation study in Section \ref{section:ablation}.

\begin{figure}
\hspace{-25pt}
    \centering
    \includegraphics[width=1\columnwidth]{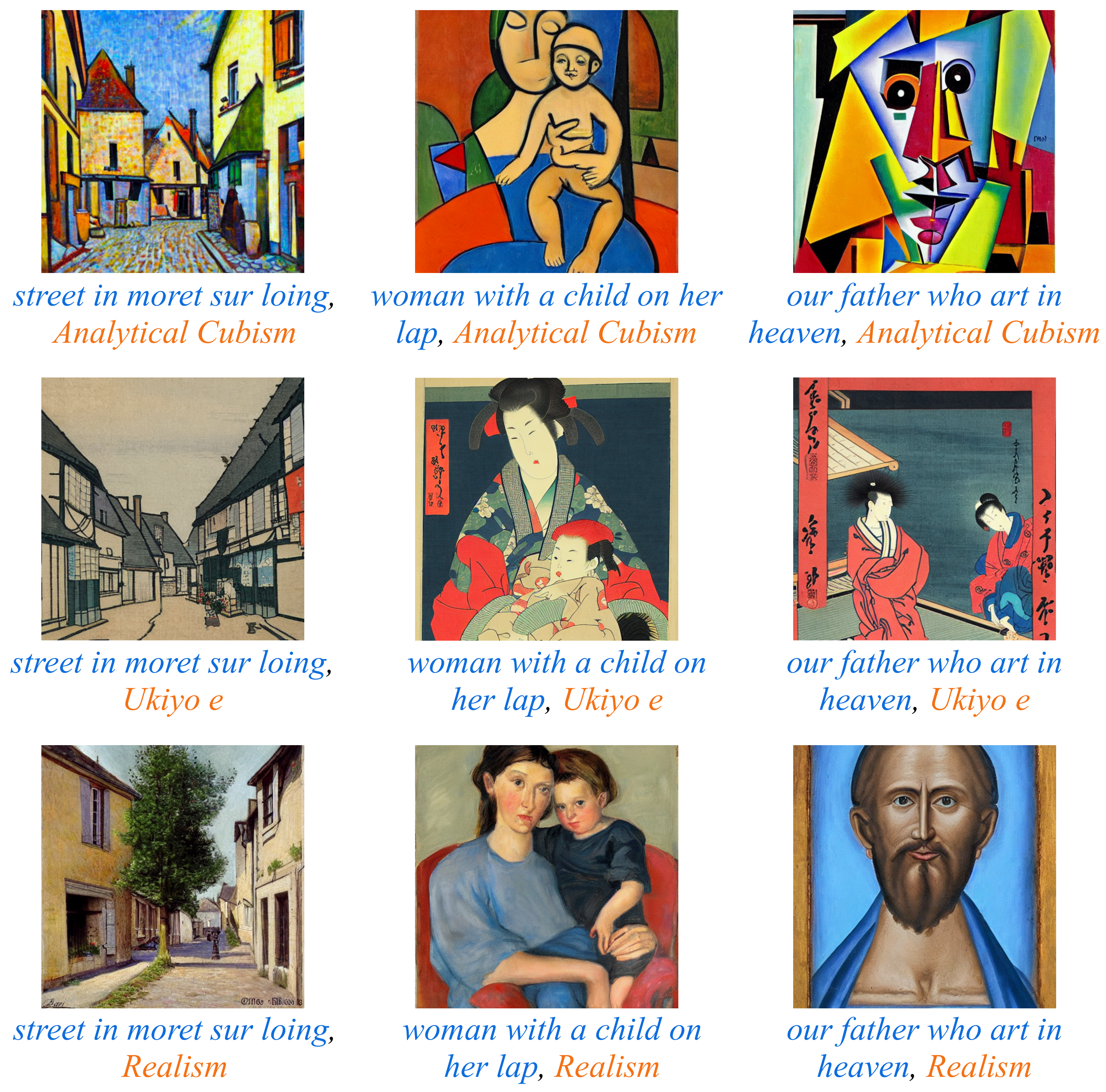}
    \vspace{-10pt}
    \caption{Examples of prompts and the corresponding generated diffusion images. The first part of the prompt (in blue) denotes the content description $x^C$, and the second part (in orange) is the style description $x^S$. Each column depicts the same content $x^C$ while each row depicts one style $x^S$.}
    \label{fig:dm}
    \vspace{-10pt}
\end{figure}

\paragraph{Evaluation data}
To evaluate content and style in the classification task, we apply genre and style movement labels in art datasets that can be served as substitutes for presenting content and style, even if they do not entirely satisfy our definition in this paper. 
In detail, the genre labels indicate the type of scene depicted in the paintings, such as ``portrait'' or ``cityscape'', while the style movement labels correspond to artistic movements such as ``Impressionism'' and ``Expressionism''. 
We use the WikiArt dataset \cite{artgan2018} for evaluation, a popular artwork dataset with both genre and style movement annotations. In total, we get $81,445$ paintings, $57,025$ in the training set, $12,210$ in the validation set, and $12,210$ in the test set, with three types of labels: $23$ artists, $10$ genres, and $27$ style movements. All evaluation results are computed on the test set.

\paragraph{Training data} 
Baselines reported on WikiArt are trained with WikiArt training set. GOYA is trained with generated images by Stable Diffusion, which are described in the next paragraph. Additionally, the training dataset of Stable Diffusion LAION-$5$B \cite{schuhmann2022laion} has over five billion image-text pairs, which contain some paintings from the WikiArt test set. We examine other models trained on generated images, which are equally affected by this issue.

\paragraph{Image generation details}
To generate images that look similar to human-made paintings, we rely on crafting prompts $x=\{x^C, x^S\}$ as explained in Section \ref{sec:StableDiffusion}. For simplicity, we choose titles of paintings as $x^C$ and style movements as $x^S$, although any other definitions of content and style descriptions can be used. In total, there are $43,610$ content descriptions $x^C$, and $27$ style descriptions $x^S$. For each $x^C$, we randomly choose five $x^S$ to generate five prompts $x$. Then, each prompt generates five images with random seeds. Altogether, we obtain $218,050$ prompts and $1,090,250$ synthetic images. We split the generated images into $981,225$ training and $109,025$ validation images. We use Stable Diffusion v1.4\footnote{\url{https://github.com/CompVis/stable-diffusion}} and generate images of size $512 \times 512$ through $50$ PLMS \cite{liu2022pseudo} sampling steps.

Figure \ref{fig:dm} shows some examples of diffusion generated images by the designed prompts. We can observe that the depicted scene is consistent with the content description in the prompts. Images in the same column have the same $x^C$ but different $x^S$, and have a high agreement in content while carrying significant differences in style. Likewise, images in the same row have the same $x^S$ but different $x^C$, and paint different scenes or objects while maintaining a similar style. 
However, some of the content descriptions are religious, such as $x^C$ in the third column, ``our father who art in heaven.'' In such cases, it may be difficult to have an agreement on the semantic consistency of the generated images and the prompts.

\paragraph{GOYA details}
\begin{table}[t]
\renewcommand{\arraystretch}{1.1}
\small
\centering
\caption{Distance Correlation (DC) between content and style embeddings on the WikiArt test set. \textit{Labels} indicate the results when using a one-hot vector embedding of the ground truth labels. ResNet50 and CLIP are fine-tuned on WikiArt while DINO loads the pre-trained weights.}
\vspace{-8pt}
\begin{tabularx}{1.005\columnwidth}{@{}l r l r r r}
\toprule
\multirow{2}{*}{Model} & Training & Training & Emb. size & Emb. size & \multirow{2}{*}{DC $\downarrow$}\\ 
& params. & data & content & style & \\
\midrule
\textit{Labels} & \textit{-} & \textit{-} & $\mathit{27}$ & $\mathit{27}$ & $\mathit{0.269}$ \\
ResNet50 \cite{He2016DeepRL} & $47\text{M}$ & WikiArt & $2,048$ & $2,048$ & $0.635$ \\
CLIP \cite{radford2021learning} & $302\text{M}$ & WikiArt & $512$ & $512$ & $0.460$ \\
DINO \cite{caron2021emerging} & $\text{-}$ & $\text{-}$ & $616,225$ & $768$ & $0.518$ \\
GOYA (Ours) & $15\text{M}$ & Diffusion & $2,048$ & $2,048$ & $\textbf{0.367}$ \\
\bottomrule
\end{tabularx}
\label{tab:dc_new}
% \vspace{-8pt}
\end{table}

\begin{table*}[t]
\renewcommand{\arraystretch}{1.1}
\setlength{\tabcolsep}{10pt}
\small
\centering
% \vspace{5pt}
\caption{Genre and style movement accuracy on WikiArt \cite{artgan2018} dataset for different models.}
% \vspace{5pt}
\begin{tabularx}{0.96\textwidth}{@{} p{1mm} l l l r r r r r}
\toprule
\multicolumn{2}{l}{\multirow{2}{*}{Model}} & Training & \multirow{2}{*}{Label} & \multirow{2}{*}{Num. train} & Emb. size & Emb. size & Accuracy & Accuracy \\
& & data & & & content & style & genre & style\\
\midrule
\rowcolor{oursrow}
\multicolumn{9}{l}{Pre-trained} \\
& Gram Matrix \cite{gatys2015neural, gatys2015texture} & - & - & - & $4,096$ & $4,096$ & $61.81$ & $40.79$ \\
& ResNet50 \cite{He2016DeepRL} & - & - & - & $2,048$ & $2,048$ & $67.85$ & $43.15$ \\
& CLIP \cite{radford2021learning} & - & - & - & $512$ & $512$ & $\textbf{71.56}$ & $\textbf{51.23}$ \\
& DINO \cite{caron2021emerging} & - & - & - & $616,225$ & $768$ & $51.13$ & $38.81$ \\

\rowcolor{oursrow}
\multicolumn{9}{l}{Trained on WikiArt} \\
& ResNet50 \cite{He2016DeepRL} (Genre) & WikiArt & Genre & $57,025$ & $2,048$ & $2,048$ & $79.13$ & $43.17$ \\
& ResNet50 \cite{He2016DeepRL} (Style) & WikiArt & Style & $57,025$ & $2,048$ & $2,048$ & $67.22$ & $\textbf{64.44}$ \\
& CLIP \cite{radford2021learning} (Genre) & WikiArt & Genre & $57,025$ & $512$ & $512$ & $\textbf{80.43}$ & $34.98$ \\
& CLIP \cite{radford2021learning} (Style) & WikiArt & Style & $57,025$ & $512$ & $512$ & $56.28$ & $63.02$ \\
& SimCLR \cite{chen2020simple} & WikiArt & - & $57,025$ & $2,048$ & $2,048$ & $65.82$ & $45.15$ \\
& SimSiam \cite{chen2021exploring} & WikiArt & -  & $57,025$ & $2,048$ & $2,048$ & $51.65$ & $31.24$ \\

\rowcolor{oursrow}
\multicolumn{9}{l}{Trained on Diffusion generated} \\
& ResNet50 \cite{He2016DeepRL} (Movement) & Diffusion & Movement & $981,225$ & $2,048$ & $2,048$ & $61.78$ & $45.79$ \\
& CLIP \cite{radford2021learning} (Movement) & Diffusion & Movement & $981,225$ & $512$ & $512$ & $52.65$ & $43.58$ \\
& SimCLR \cite{chen2020simple} & Diffusion & - & $981,225$ & $2,048$ & $2,048$ & $33.82$ & $20.88$ \\
& GOYA (Ours) & Diffusion & - & $981,225$ & $2,048$ & $2,048$ & $\textbf{69.70}$ & $\textbf{50.90}$ \\

\bottomrule
\end{tabularx}
% \vspace{5pt}
\label{tab:clf_new}
\end{table*}

For the CLIP image and text encoders, we load the pre-trained weights of CLIP-ViT-B/32 models.\footnote{\url{https://github.com/openai/CLIP}} The margin for computing contrastive losses $\epsilon^C=\epsilon^S=0.5$. In the indicator function for the content contrastive loss, the threshold $\epsilon^T$ is set to $0.25$.
We use Adam optimizer \cite{kingma2014adam} with base learning rate $= 0.0005$ and decay rate $= 0.9$. We train GOYA on $4$ A$6000$ GPUs with Distributed Data Parallel in PyTorch.\footnote{\url{https://pytorch.org/}} In each device, the batch size is set as $512$. Before feeding into CLIP, images are resized to $224\times 224$ pixels. 

% ----------------------------------------------------------------------------
\subsection{Disentanglement evaluation} \label{section:DC}
To measure content and style disentanglement quantitatively, we compute the Distance Correlation (DC) \cite{liu2020measuring} between content and style embeddings, which is specially designed for content and style disentanglement evaluation. Let $G^C$ and $G^S$ denote matrices containing all content and style embeddings in the WikiArt test set, i.e., $G^C = (g^C_1\;\;\cdots\;\;g^C_N)$ and $G^S = (g^S_1 \cdots \;\; g^S_N)$. For an arbitrary pair $(i, j)$ of embeddings, the distances $p^C_{ij}$ and $q^S_{ij}$ can be computed by:
\begin{equation}
    p^C_{ij} = \|g^C_i - g^C_j\|, \;\;\;
    p^S_{ij} = \|g^S_i - g^S_j\|,
\end{equation}
where $\|\cdot\|$ gives the Euclidean distance. Let $\bar{p}^C_{i\cdot}$, $\bar{p}^C_{\cdot j}$, and $\bar{p}^C$ denote the means over $j$, $i$, and both $i$ and $j$, respectively. With these means, the distances can be doubly centered by
\begin{equation}
q^C_{ij} = p^C_{ij} - \bar{p}^C_{i\cdot} - \bar{p}^C_{\cdot j} + \bar{p}^C,
\end{equation}
and likewise for $q^S_{ij}$. DC between $G^C$ and $G^S$ is given by:
\begin{equation}
    \text{DC}(G^C, G^S) = \frac{\text{dCov}(G^C, G^S)}{\sqrt{\text{dCov}(G^C, G^C)\text{dCov}(G^S, G^S)}},
\end{equation}
where 
\begin{equation}
    \text{dCov}(G^C, G^S) = \frac{1}{N} \sqrt{\sum\nolimits_i \sum\nolimits_j q^C_{ij} q^S_{ij}}.
\end{equation}
$\text{dCov}(G^C, G^C)$ and $\text{dCov}(G^S, G^S)$ are defined likewise. DC can be computed for arbitrary matrices with $N$ columns. $\text{DC}$ is in $[0, 1]$, and lower value means $G^C$ and $G^{S}$ are less correlated. We aim at DC being close to 0. 

\paragraph{Baselines}
To compute the lower bound DC on the WikiArt test dataset, we assign the one-hot vector of the ground-truth genre and style movement labels as the content and style embeddings, representing the uppermost disentanglement when the labels are $100\%$ correct. Besides the lower bound, we evaluate DC on ResNet50 \cite{He2016DeepRL}, CLIP \cite{radford2021learning} and DINO \cite{caron2021emerging}. 
For ResNet50, embeddings are extracted before the last fully-connected layer. For CLIP, we use the embedding from the CLIP image encoder $\mathcal{E}_I$. For pre-trained DINO, following Splice \cite{tumanyan2022splicing}, content and style embeddings are extracted at the deepest layer from the self-similarity of keys in the attention module and the $[\mathrm{CLS}]$ token, respectively. 

\paragraph{Results}
Results are reported in Table \ref{tab:dc_new}. GOYA shows the best disentanglement with the lowest DC, $0.367$, and a large margin with the second-best disentangled embeddings from fine-tuned CLIP. With only nearly $1/3$ training parameters of ResNet50 and $1/20$ of CLIP, GOYA outperforms embeddings directly trained on WikiArt's real paintings while consuming fewer resources. Also, GOYA achieves better disentanglement capability than DINO, with much more compact embeddings, e.g. $1/300$ content size embedding. However, there is still a notorious gap between GOYA and the lower bound based on labels, showing that there is room for improvement. 

\begin{figure*}
    \centering
    \includegraphics[width=0.68\textwidth]{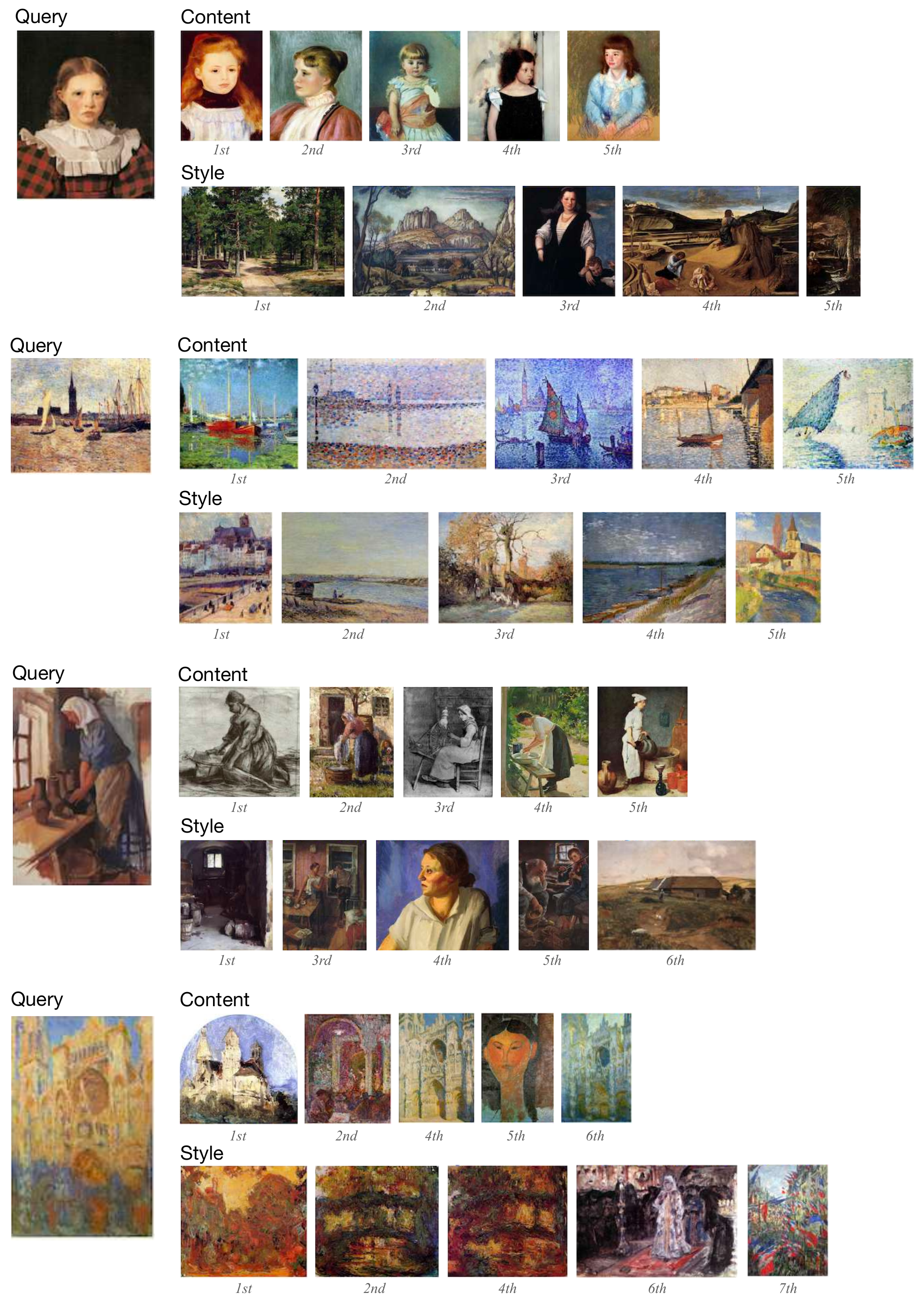}
    \vspace{-3pt}
    \caption{Retrieval results on the WikiArt test set based on cosine similarity. The similarity decreases from left to right. Copyrighted images are skipped.
    }
    \label{fig:similarity}
    \vspace{-10pt}
\end{figure*}

% ----------------------------------------------------------------------------
\subsection{Classification evaluation} \label{section:classification}

For evaluating the disentangled embeddings for art classification, following the protocol in \cite{chen2021exploring}, we train two independent classifiers with a single linear layer on top of the content and style embeddings. 

\paragraph{Baselines}
We compare GOYA against three types of baselines: pre-trained models, models trained on WikiArt dataset, and models trained on diffusion generated images. As pre-trained models, we use Gram matrix \cite{gatys2015neural, gatys2015texture}, ResNet50 \cite{He2016DeepRL}, CLIP \cite{radford2021learning} and DINO \cite{caron2021emerging}. 
For models trained on WikiArt, other than fine-tuning ResNet50 and CLIP, we also apply two popular contrastive learning methods: SimCLR \cite{chen2020simple} and SimSiam \cite{chen2021exploring}. 
For models trained on generated images, ResNet50 and CLIP are fine-tuned with style movements in the prompts. SimCLR is trained without any annotations.

\paragraph{Results}

Table \ref{tab:clf_new} shows the classification results. Compared with the pre-trained baselines in the first four rows, GOYA is ahead of Gram matrix, ResNet50, and DINO. Yet, it lags behind pre-trained CLIP by less than $1\%$ in both genre and style movement accuracy. 
Compared with models trained on WikiArt, although not comparable to fine-tuned ResNet50 and CLIP on classification, GOYA achieves better capability of disentanglement as shown in Table \ref{tab:dc_new}.
Also, GOYA enables better performance on classification against contrastive learning models SimCLR and SimSiam.

When training on diffusion generated images, GOYA achieves the best classification performance compared to other models with different embedding sizes. After fine-tuned on style movement in the prompts, ResNet50 increases $3\%$ on the style accuracy, showing the potential for analysis via synthetically generated images. However, CLIP decreases in both genre and style accuracy after fine-tuning on generated images. SimCLR has a dramatic decrement when trained on generated images compared to on WikiArt. As SimCLR focuses more on learning the intricacies of the image itself rather than the relation of images, it learns the distribution of generated images, leading to poor performance on WikiArt. While training on the same dataset, GOYA sustains better capability on classification tasks while achieving high disentanglement.

% ----------------------------------------------------------------------------
\subsection{Similarity retrieval} \label{section:similarity}
Next, we evaluate the visual retrieval performance of GOYA.
Given a painting as a query, the five closest images are retrieved based on the cosine similarity of the embeddings in the content and style space, representing the most similar paintings in each space.

\paragraph{Results}
Visual results are shown in Figure \ref{fig:similarity}. Most of the paintings retrieved in the content space depict scenes similar to the query image. For instance, in the third query image, there is a woman with a headscarf bending over to scrub a pot, while all similar paintings in the content space show a woman leaning to do manual labor such as washing, knitting, and chopping, independently of their visual style. It can be seen that in most similar content paintings, various styles are depicted through different color compositions and tones. On the contrary, similar paintings in the style space are prone to carry similar styles but different content. The similar style images of the query image have similar color compositions or brushstrokes, but depict different scenes compared to the query image. 
For example, the fourth query image, which is one of the paintings in the ``\textit{Rouen Cathedral}'' series by Monet, exhibits different visual appearances on the same object under the light variance. It can be observed that the retrieved images in the style space also apply a different light condition to create a sense of space and display vivid color contrast. Not only that, but they also display similar color compositions and strokes, but paint different scenes.

% ----------------------------------------------------------------------------
\subsection{Ablation study} \label{section:ablation}
We conduct an ablation study on WikiArt test set to examine the effectiveness of the losses and the network structure in GOYA.

\paragraph{Losses}
We compare the losses in GOYA against two other popular contrastive losses, Triplet loss \cite{schroff2015facenet} and NTXent loss \cite{sohn2016improved}, both of which have shown their superiority in many contrastive learning methods. We also investigate applying style classification loss on top of the above-mentioned contrastive losses. The selection criteria of positive and negative pairs are the same for all the losses. 

\begin{figure}
\hspace{-15pt}
    \centering
    \includegraphics[width=0.92\columnwidth]{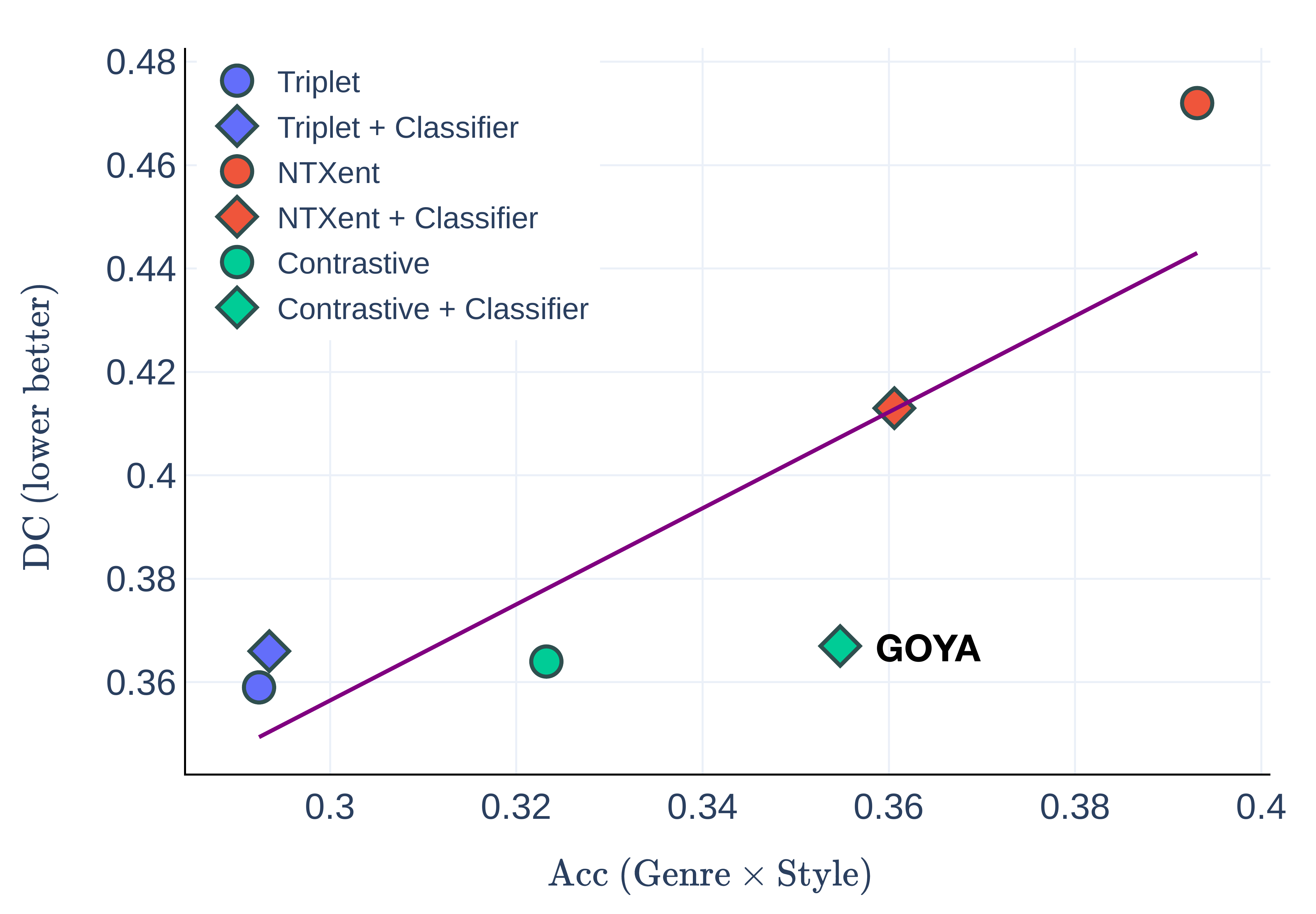}
    \vspace{-5pt}
    \caption{Loss comparison. The $x$-axis shows the product of genre and style accuracies (the higher the better) while the $y$-axis presents the disentanglement, DC (the lower the better). The purple line shows the trendline as $y =  0.0776 + 0.9295x$. In general, better accuracy is obtained at expense of a worse disentanglement. Only GOYA (Contrastive + Classifier loss) improves accuracy without damaging DC.}
    \label{fig:ablation}
    \vspace{-10pt}
\end{figure}

The results in terms of accuracy (as the product of genre and style movement accuracies) and disentanglement (as DC) are shown in Figure \ref{fig:ablation}. 
The NTXent loss achieves the highest accuracy but with the cost of undercutting disentanglement ability. In contrast, triplet loss has almost the best disentanglement performance but is at a disadvantage in terms of classification performance. Compared to those two losses, only contrastive loss in GOYA is able to sustain a balance between disentanglement and classification performance. Moreover, after occupying the classification loss, GOYA has a boost in classification without sacrificing disentanglement, achieving the best performance compared to the other loss settings. 

\begin{figure}
\hspace{-10pt}
    \centering
    \includegraphics[width=0.92\columnwidth]{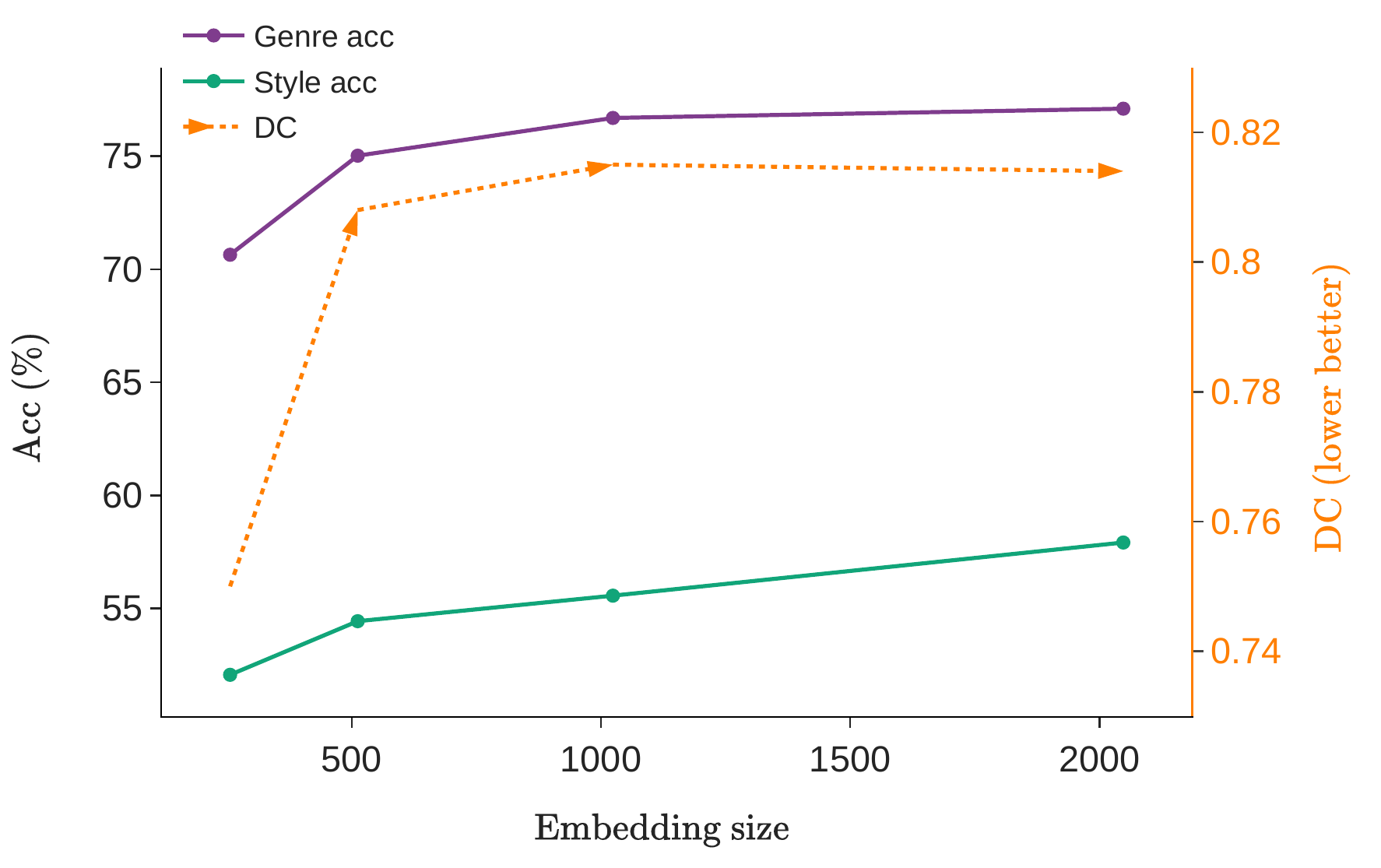}
    \vspace{-5pt}
    \caption{Disentanglement and classification evaluation with different embedding sizes when only one single layer is set in the content and style encoder.}
    \label{fig:ablation_layer}
    \vspace{-20pt}
\end{figure}

\paragraph{Embedding size}
We explore the effect of the embedding size on a single layer content and style encoders, with embedding sizes ranging from $256$ to $2,048$. Figure \ref{fig:ablation_layer} shows that the accuracy of both genre and style improves up to $6\%$ as the embedding size increases, but conversely, the DC becomes worse, from $0.750$ to $0.814$, indicating a trade-off between classification and disentanglement. Moreover, the classification performance of genre and style movement outperforms the pre-trained CLIP (shown in Table \ref{tab:clf_new}) when the embedding size exceeds $512$, suggesting that larger embedding sizes carry a stronger ability to distill knowledge from the pre-trained model. Inspired by this finding, we set the embedding size to $2,048$.

\section{Conclusion}
This work proposed GOYA, a method for disentangling content and style embeddings of paintings by training on synthetic images generated with Stable Diffusion. Exploiting the multi-modal CLIP latent space, we first extracted off-the-shelf embeddings to then learn similarities and dissimilarities in content and style with two encoders trained with contrastive learning. Evaluation on the WikiArt dataset included disentanglement, classification, and similarity retrieval. Despite relying only on synthetic images, results showed that GOYA achieves good disentanglement between content and style embeddings. We this work sheds light on the adoption of generative models in the analysis of the digital humanities. 

% \footnote{This work is partly supported by JST FOREST Grant No. JPMJFR216O and JSPS KAKENHI Grant No. 23H00497 and No. 20K19822.}

\begin{acks}
This work is partly supported by JST FOREST Grant No. JPMJFR216O and JSPS KAKENHI Grant No. 23H00497 and No. 20K19822.
\end{acks}

\bibliographystyle{ACM-Reference-Format}
\bibliography{acmart}


\begin{thebibliography}{88}

%%% the Web address.

\ifx \showCODEN    \undefined \def \showCODEN     #1{\unskip}     \fi
\ifx \showDOI      \undefined \def \showDOI       #1{#1}\fi
\ifx \showISBNx    \undefined \def \showISBNx     #1{\unskip}     \fi
\ifx \showISBNxiii \undefined \def \showISBNxiii  #1{\unskip}     \fi
\ifx \showISSN     \undefined \def \showISSN      #1{\unskip}     \fi
\ifx \showLCCN     \undefined \def \showLCCN      #1{\unskip}     \fi
\ifx \shownote     \undefined \def \shownote      #1{#1}          \fi
\ifx \showarticletitle \undefined \def \showarticletitle #1{#1}   \fi
\ifx \showURL      \undefined \def \showURL       {\relax}        \fi
% invisible to TeX
\providecommand\bibfield[2]{#2}
\providecommand\bibinfo[2]{#2}
\providecommand\natexlab[1]{#1}
\providecommand\showeprint[2][]{arXiv:#2}

\bibitem[Bai et~al\mbox{.}(2021)]%
        {bai2021explain}
\bibfield{author}{\bibinfo{person}{Zechen Bai}, \bibinfo{person}{Yuta
  Nakashima}, {and} \bibinfo{person}{Noa Garcia}.}
  \bibinfo{year}{2021}\natexlab{}.
\newblock \showarticletitle{Explain me the painting: Multi-topic knowledgeable
  art description generation}. In \bibinfo{booktitle}{\emph{ICCV}}.
  \bibinfo{pages}{5422--5432}.
\newblock


\bibitem[Carneiro et~al\mbox{.}(2012)]%
        {carneiro2012artistic}
\bibfield{author}{\bibinfo{person}{Gustavo Carneiro}, \bibinfo{person}{Nuno
  Pinho~da Silva}, \bibinfo{person}{Alessio~Del Bue}, {and}
  \bibinfo{person}{Jo{\~a}o~Paulo Costeira}.} \bibinfo{year}{2012}\natexlab{}.
\newblock \showarticletitle{Artistic image classification: An analysis on the
  printart database}. In \bibinfo{booktitle}{\emph{ECCV}}. Springer,
  \bibinfo{pages}{143--157}.
\newblock


\bibitem[Caron et~al\mbox{.}(2021)]%
        {caron2021emerging}
\bibfield{author}{\bibinfo{person}{Mathilde Caron}, \bibinfo{person}{Hugo
  Touvron}, \bibinfo{person}{Ishan Misra}, \bibinfo{person}{Herv{\'e}
  J{\'e}gou}, \bibinfo{person}{Julien Mairal}, \bibinfo{person}{Piotr
  Bojanowski}, {and} \bibinfo{person}{Armand Joulin}.}
  \bibinfo{year}{2021}\natexlab{}.
\newblock \showarticletitle{Emerging properties in self-supervised vision
  transformers}. In \bibinfo{booktitle}{\emph{ICCV}}.
  \bibinfo{pages}{9650--9660}.
\newblock


\bibitem[Cetinic(2021)]%
        {cetinic2021towards}
\bibfield{author}{\bibinfo{person}{Eva Cetinic}.}
  \bibinfo{year}{2021}\natexlab{}.
\newblock \showarticletitle{Towards generating and evaluating iconographic
  image captions of artworks}.
\newblock \bibinfo{journal}{\emph{Journal of Imaging}} \bibinfo{volume}{7},
  \bibinfo{number}{8} (\bibinfo{year}{2021}), \bibinfo{pages}{123}.
\newblock


\bibitem[Cetinic et~al\mbox{.}(2018)]%
        {cetinic2018fine}
\bibfield{author}{\bibinfo{person}{Eva Cetinic}, \bibinfo{person}{Tomislav
  Lipic}, {and} \bibinfo{person}{Sonja Grgic}.}
  \bibinfo{year}{2018}\natexlab{}.
\newblock \showarticletitle{Fine-tuning convolutional neural networks for fine
  art classification}.
\newblock \bibinfo{journal}{\emph{Expert Systems with Applications}}
  \bibinfo{volume}{114} (\bibinfo{year}{2018}), \bibinfo{pages}{107--118}.
\newblock


\bibitem[Chen et~al\mbox{.}(2021a)]%
        {chen2021dualast}
\bibfield{author}{\bibinfo{person}{Haibo Chen}, \bibinfo{person}{Lei Zhao},
  \bibinfo{person}{Zhizhong Wang}, \bibinfo{person}{Huiming Zhang},
  \bibinfo{person}{Zhiwen Zuo}, \bibinfo{person}{Ailin Li},
  \bibinfo{person}{Wei Xing}, {and} \bibinfo{person}{Dongming Lu}.}
  \bibinfo{year}{2021}\natexlab{a}.
\newblock \showarticletitle{Dualast: Dual style-learning networks for artistic
  style transfer}. In \bibinfo{booktitle}{\emph{CVPR}}.
  \bibinfo{pages}{872--881}.
\newblock


\bibitem[Chen et~al\mbox{.}(2021b)]%
        {chen2021diverse}
\bibfield{author}{\bibinfo{person}{Haibo Chen}, \bibinfo{person}{Lei Zhao},
  \bibinfo{person}{Huiming Zhang}, \bibinfo{person}{Zhizhong Wang},
  \bibinfo{person}{Zhiwen Zuo}, \bibinfo{person}{Ailin Li},
  \bibinfo{person}{Wei Xing}, {and} \bibinfo{person}{Dongming Lu}.}
  \bibinfo{year}{2021}\natexlab{b}.
\newblock \showarticletitle{Diverse image style transfer via invertible
  cross-space mapping}. In \bibinfo{booktitle}{\emph{ICCV}}. IEEE Computer
  Society, \bibinfo{pages}{14860--14869}.
\newblock


\bibitem[Chen and Yang(2019)]%
        {chen2019recognizing}
\bibfield{author}{\bibinfo{person}{Liyi Chen} {and} \bibinfo{person}{Jufeng
  Yang}.} \bibinfo{year}{2019}\natexlab{}.
\newblock \showarticletitle{Recognizing the style of visual arts via adaptive
  cross-layer correlation}. In \bibinfo{booktitle}{\emph{ACM MM}}.
  \bibinfo{pages}{2459--2467}.
\newblock


\bibitem[Chen et~al\mbox{.}(2020)]%
        {chen2020simple}
\bibfield{author}{\bibinfo{person}{Ting Chen}, \bibinfo{person}{Simon
  Kornblith}, \bibinfo{person}{Mohammad Norouzi}, {and}
  \bibinfo{person}{Geoffrey Hinton}.} \bibinfo{year}{2020}\natexlab{}.
\newblock \showarticletitle{A simple framework for contrastive learning of
  visual representations}. In \bibinfo{booktitle}{\emph{ICML}}. PMLR,
  \bibinfo{pages}{1597--1607}.
\newblock


\bibitem[Chen et~al\mbox{.}(2016)]%
        {chen2016infogan}
\bibfield{author}{\bibinfo{person}{Xi Chen}, \bibinfo{person}{Yan Duan},
  \bibinfo{person}{Rein Houthooft}, \bibinfo{person}{John Schulman},
  \bibinfo{person}{Ilya Sutskever}, {and} \bibinfo{person}{Pieter Abbeel}.}
  \bibinfo{year}{2016}\natexlab{}.
\newblock \showarticletitle{Info{GAN}: Interpretable representation learning by
  information maximizing generative adversarial nets}.
\newblock \bibinfo{journal}{\emph{NeurIPS}}  \bibinfo{volume}{29}
  (\bibinfo{year}{2016}).
\newblock


\bibitem[Chen and He(2021)]%
        {chen2021exploring}
\bibfield{author}{\bibinfo{person}{Xinlei Chen} {and} \bibinfo{person}{Kaiming
  He}.} \bibinfo{year}{2021}\natexlab{}.
\newblock \showarticletitle{Exploring simple siamese representation learning}.
  In \bibinfo{booktitle}{\emph{CVPR}}. \bibinfo{pages}{15750--15758}.
\newblock


\bibitem[Cheng et~al\mbox{.}(2021)]%
        {cheng2021data}
\bibfield{author}{\bibinfo{person}{Ruizhe Cheng}, \bibinfo{person}{Bichen Wu},
  \bibinfo{person}{Peizhao Zhang}, \bibinfo{person}{Peter Vajda}, {and}
  \bibinfo{person}{Joseph~E Gonzalez}.} \bibinfo{year}{2021}\natexlab{}.
\newblock \showarticletitle{Data-efficient language-supervised zero-shot
  learning with self-distillation}. In \bibinfo{booktitle}{\emph{CVPR}}.
  \bibinfo{pages}{3119--3124}.
\newblock


\bibitem[Chu and Wu(2018)]%
        {chu2018image}
\bibfield{author}{\bibinfo{person}{Wei-Ta Chu} {and} \bibinfo{person}{Yi-Ling
  Wu}.} \bibinfo{year}{2018}\natexlab{}.
\newblock \showarticletitle{Image style classification based on learnt deep
  correlation features}.
\newblock \bibinfo{journal}{\emph{Transactions on Multimedia}}
  \bibinfo{volume}{20}, \bibinfo{number}{9} (\bibinfo{year}{2018}),
  \bibinfo{pages}{2491--2502}.
\newblock


\bibitem[Collomosse et~al\mbox{.}(2017)]%
        {collomosse2017sketching}
\bibfield{author}{\bibinfo{person}{John Collomosse}, \bibinfo{person}{Tu Bui},
  \bibinfo{person}{Michael~J Wilber}, \bibinfo{person}{Chen Fang}, {and}
  \bibinfo{person}{Hailin Jin}.} \bibinfo{year}{2017}\natexlab{}.
\newblock \showarticletitle{Sketching with style: Visual search with sketches
  and aesthetic context}. In \bibinfo{booktitle}{\emph{ICCV}}.
  \bibinfo{pages}{2660--2668}.
\newblock


\bibitem[Crowley and Zisserman(2014)]%
        {crowley2014state}
\bibfield{author}{\bibinfo{person}{Elliot~J Crowley} {and}
  \bibinfo{person}{Andrew Zisserman}.} \bibinfo{year}{2014}\natexlab{}.
\newblock \showarticletitle{The state of the art: Object retrieval in paintings
  using discriminative regions}. In \bibinfo{booktitle}{\emph{BMVC}}.
\newblock


\bibitem[Deng et~al\mbox{.}(2020)]%
        {deng2020arbitrary}
\bibfield{author}{\bibinfo{person}{Yingying Deng}, \bibinfo{person}{Fan Tang},
  \bibinfo{person}{Weiming Dong}, \bibinfo{person}{Wen Sun},
  \bibinfo{person}{Feiyue Huang}, {and} \bibinfo{person}{Changsheng Xu}.}
  \bibinfo{year}{2020}\natexlab{}.
\newblock \showarticletitle{Arbitrary style transfer via multi-adaptation
  network}. In \bibinfo{booktitle}{\emph{ACM MM}}. \bibinfo{pages}{2719--2727}.
\newblock


\bibitem[Denton et~al\mbox{.}(2017)]%
        {denton2017unsupervised}
\bibfield{author}{\bibinfo{person}{Emily~L Denton} {et~al\mbox{.}}}
  \bibinfo{year}{2017}\natexlab{}.
\newblock \showarticletitle{Unsupervised learning of disentangled
  representations from video}.
\newblock \bibinfo{journal}{\emph{NeurIPS}}  \bibinfo{volume}{30}
  (\bibinfo{year}{2017}).
\newblock


\bibitem[Ding et~al\mbox{.}(2021)]%
        {ding2021cogview}
\bibfield{author}{\bibinfo{person}{Ming Ding}, \bibinfo{person}{Zhuoyi Yang},
  \bibinfo{person}{Wenyi Hong}, \bibinfo{person}{Wendi Zheng},
  \bibinfo{person}{Chang Zhou}, \bibinfo{person}{Da Yin},
  \bibinfo{person}{Junyang Lin}, \bibinfo{person}{Xu Zou},
  \bibinfo{person}{Zhou Shao}, \bibinfo{person}{Hongxia Yang}, {et~al\mbox{.}}}
  \bibinfo{year}{2021}\natexlab{}.
\newblock \showarticletitle{Cog{V}iew: Mastering text-to-image generation via
  transformers}.
\newblock \bibinfo{journal}{\emph{NeurIPS}}  \bibinfo{volume}{34}
  (\bibinfo{year}{2021}), \bibinfo{pages}{19822--19835}.
\newblock


\bibitem[Dosovitskiy et~al\mbox{.}(2021)]%
        {dosovitskiy2020image}
\bibfield{author}{\bibinfo{person}{Alexey Dosovitskiy}, \bibinfo{person}{Lucas
  Beyer}, \bibinfo{person}{Alexander Kolesnikov}, \bibinfo{person}{Dirk
  Weissenborn}, \bibinfo{person}{Xiaohua Zhai}, \bibinfo{person}{Thomas
  Unterthiner}, \bibinfo{person}{Mostafa Dehghani}, \bibinfo{person}{Matthias
  Minderer}, \bibinfo{person}{Georg Heigold}, \bibinfo{person}{Sylvain Gelly},
  {et~al\mbox{.}}} \bibinfo{year}{2021}\natexlab{}.
\newblock \showarticletitle{An image is worth 16x16 words: {T}ransformers for
  image recognition at scale}.
\newblock \bibinfo{journal}{\emph{ICLR}} (\bibinfo{year}{2021}).
\newblock


\bibitem[El~Vaigh et~al\mbox{.}(2021)]%
        {el2021gcnboost}
\bibfield{author}{\bibinfo{person}{Cheikh~Brahim El~Vaigh},
  \bibinfo{person}{Noa Garcia}, \bibinfo{person}{Benjamin Renoust},
  \bibinfo{person}{Chenhui Chu}, \bibinfo{person}{Yuta Nakashima}, {and}
  \bibinfo{person}{Hajime Nagahara}.} \bibinfo{year}{2021}\natexlab{}.
\newblock \showarticletitle{{GCNB}oost: Artwork classification by label
  propagation through a knowledge graph}. In \bibinfo{booktitle}{\emph{ICMR}}.
  \bibinfo{pages}{92--100}.
\newblock


\bibitem[Elgammal et~al\mbox{.}(2018)]%
        {elgammal2018shape}
\bibfield{author}{\bibinfo{person}{Ahmed Elgammal}, \bibinfo{person}{Bingchen
  Liu}, \bibinfo{person}{Diana Kim}, \bibinfo{person}{Mohamed Elhoseiny}, {and}
  \bibinfo{person}{Marian Mazzone}.} \bibinfo{year}{2018}\natexlab{}.
\newblock \showarticletitle{The shape of art history in the eyes of the
  machine}. In \bibinfo{booktitle}{\emph{AAAI}}, Vol.~\bibinfo{volume}{32}.
\newblock


\bibitem[Florea et~al\mbox{.}(2016)]%
        {florea2016pandora}
\bibfield{author}{\bibinfo{person}{Corneliu Florea},
  \bibinfo{person}{R{\u{a}}zvan Condorovici}, \bibinfo{person}{Constantin
  Vertan}, \bibinfo{person}{Raluca Butnaru}, \bibinfo{person}{Laura Florea},
  {and} \bibinfo{person}{Ruxandra Vr{\^a}nceanu}.}
  \bibinfo{year}{2016}\natexlab{}.
\newblock \showarticletitle{Pandora: Description of a painting database for art
  movement recognition with baselines and perspectives}. In
  \bibinfo{booktitle}{\emph{European Signal Processing Conference}}.
  \bibinfo{pages}{918--922}.
\newblock


\bibitem[Gabbay and Hoshen(2020)]%
        {gabbay2020improving}
\bibfield{author}{\bibinfo{person}{Aviv Gabbay} {and} \bibinfo{person}{Yedid
  Hoshen}.} \bibinfo{year}{2020}\natexlab{}.
\newblock \showarticletitle{Improving style-content disentanglement in
  image-to-image translation}.
\newblock \bibinfo{journal}{\emph{arXiv preprint arXiv:2007.04964}}
  (\bibinfo{year}{2020}).
\newblock


\bibitem[Garcia et~al\mbox{.}(2019)]%
        {garcia2019context}
\bibfield{author}{\bibinfo{person}{Noa Garcia}, \bibinfo{person}{Benjamin
  Renoust}, {and} \bibinfo{person}{Yuta Nakashima}.}
  \bibinfo{year}{2019}\natexlab{}.
\newblock \showarticletitle{Context-aware embeddings for automatic art
  analysis}. In \bibinfo{booktitle}{\emph{ICMR}}. \bibinfo{pages}{25--33}.
\newblock


\bibitem[Garcia et~al\mbox{.}(2020)]%
        {garcia2020contextnet}
\bibfield{author}{\bibinfo{person}{Noa Garcia}, \bibinfo{person}{Benjamin
  Renoust}, {and} \bibinfo{person}{Yuta Nakashima}.}
  \bibinfo{year}{2020}\natexlab{}.
\newblock \showarticletitle{Context{N}et: representation and exploration for
  painting classification and retrieval in context}.
\newblock \bibinfo{journal}{\emph{International Journal of Multimedia
  Information Retrieval}} \bibinfo{volume}{9}, \bibinfo{number}{1}
  (\bibinfo{year}{2020}), \bibinfo{pages}{17--30}.
\newblock


\bibitem[Garcia and Vogiatzis(2018)]%
        {garcia2018read}
\bibfield{author}{\bibinfo{person}{Noa Garcia} {and} \bibinfo{person}{George
  Vogiatzis}.} \bibinfo{year}{2018}\natexlab{}.
\newblock \showarticletitle{How to read paintings: semantic art understanding
  with multi-modal retrieval}. In \bibinfo{booktitle}{\emph{ECCV Workshops}}.
  \bibinfo{pages}{0--0}.
\newblock


\bibitem[Gatys et~al\mbox{.}(2015a)]%
        {gatys2015neural}
\bibfield{author}{\bibinfo{person}{LA Gatys}, \bibinfo{person}{AS Ecker}, {and}
  \bibinfo{person}{M Bethge}.} \bibinfo{year}{2015}\natexlab{a}.
\newblock \showarticletitle{A Neural Algorithm of Artistic Style}.
\newblock \bibinfo{journal}{\emph{Nature Communications}}
  (\bibinfo{year}{2015}).
\newblock


\bibitem[Gatys et~al\mbox{.}(2015b)]%
        {gatys2015texture}
\bibfield{author}{\bibinfo{person}{Leon Gatys}, \bibinfo{person}{Alexander~S
  Ecker}, {and} \bibinfo{person}{Matthias Bethge}.}
  \bibinfo{year}{2015}\natexlab{b}.
\newblock \showarticletitle{Texture synthesis using convolutional neural
  networks}.
\newblock \bibinfo{journal}{\emph{NeurIPS}}  \bibinfo{volume}{28}
  (\bibinfo{year}{2015}).
\newblock


\bibitem[Gatys et~al\mbox{.}(2016)]%
        {gatys2016image}
\bibfield{author}{\bibinfo{person}{Leon~A Gatys}, \bibinfo{person}{Alexander~S
  Ecker}, {and} \bibinfo{person}{Matthias Bethge}.}
  \bibinfo{year}{2016}\natexlab{}.
\newblock \showarticletitle{Image style transfer using convolutional neural
  networks}. In \bibinfo{booktitle}{\emph{CVPR}}. \bibinfo{pages}{2414--2423}.
\newblock


\bibitem[Gonthier et~al\mbox{.}(2018)]%
        {gonthier2018weakly}
\bibfield{author}{\bibinfo{person}{Nicolas Gonthier}, \bibinfo{person}{Yann
  Gousseau}, \bibinfo{person}{Said Ladjal}, {and} \bibinfo{person}{Olivier
  Bonfait}.} \bibinfo{year}{2018}\natexlab{}.
\newblock \showarticletitle{Weakly Supervised Object Detection in Artworks}. In
  \bibinfo{booktitle}{\emph{ECCV Workshops}}. \bibinfo{pages}{692--709}.
\newblock


\bibitem[Gonzalez-Garcia et~al\mbox{.}(2018)]%
        {gonzalez2018image}
\bibfield{author}{\bibinfo{person}{Abel Gonzalez-Garcia},
  \bibinfo{person}{Joost Van De~Weijer}, {and} \bibinfo{person}{Yoshua
  Bengio}.} \bibinfo{year}{2018}\natexlab{}.
\newblock \showarticletitle{Image-to-image translation for cross-domain
  disentanglement}.
\newblock \bibinfo{journal}{\emph{NeurIPS}}  \bibinfo{volume}{31}
  (\bibinfo{year}{2018}).
\newblock


\bibitem[He et~al\mbox{.}(2016)]%
        {He2016DeepRL}
\bibfield{author}{\bibinfo{person}{Kaiming He}, \bibinfo{person}{X. Zhang},
  \bibinfo{person}{Shaoqing Ren}, {and} \bibinfo{person}{Jian Sun}.}
  \bibinfo{year}{2016}\natexlab{}.
\newblock \showarticletitle{Deep Residual Learning for Image Recognition}.
\newblock \bibinfo{journal}{\emph{CVPR}} (\bibinfo{year}{2016}),
  \bibinfo{pages}{770--778}.
\newblock


\bibitem[Higgins et~al\mbox{.}(2017)]%
        {higgins2016beta}
\bibfield{author}{\bibinfo{person}{Irina Higgins}, \bibinfo{person}{Loic
  Matthey}, \bibinfo{person}{Arka Pal}, \bibinfo{person}{Christopher Burgess},
  \bibinfo{person}{Xavier Glorot}, \bibinfo{person}{Matthew Botvinick},
  \bibinfo{person}{Shakir Mohamed}, {and} \bibinfo{person}{Alexander
  Lerchner}.} \bibinfo{year}{2017}\natexlab{}.
\newblock \showarticletitle{$\beta$-{VAE}: Learning basic visual concepts with
  a constrained variational framework}. In \bibinfo{booktitle}{\emph{ICLR}}.
\newblock


\bibitem[Ho et~al\mbox{.}(2020)]%
        {ho2020denoising}
\bibfield{author}{\bibinfo{person}{Jonathan Ho}, \bibinfo{person}{Ajay Jain},
  {and} \bibinfo{person}{Pieter Abbeel}.} \bibinfo{year}{2020}\natexlab{}.
\newblock \showarticletitle{Denoising diffusion probabilistic models}.
\newblock \bibinfo{journal}{\emph{NeurIPS}}  \bibinfo{volume}{33}
  (\bibinfo{year}{2020}), \bibinfo{pages}{6840--6851}.
\newblock


\bibitem[Huang and Belongie(2017)]%
        {huang2017adain}
\bibfield{author}{\bibinfo{person}{Xun Huang} {and} \bibinfo{person}{Serge
  Belongie}.} \bibinfo{year}{2017}\natexlab{}.
\newblock \showarticletitle{Arbitrary Style Transfer in Real-time with Adaptive
  Instance Normalization}. In \bibinfo{booktitle}{\emph{ICCV}}.
\newblock


\bibitem[Huang et~al\mbox{.}(2018)]%
        {huang2018multimodal}
\bibfield{author}{\bibinfo{person}{Xun Huang}, \bibinfo{person}{Ming-Yu Liu},
  \bibinfo{person}{Serge Belongie}, {and} \bibinfo{person}{Jan Kautz}.}
  \bibinfo{year}{2018}\natexlab{}.
\newblock \showarticletitle{Multimodal unsupervised image-to-image
  translation}. In \bibinfo{booktitle}{\emph{ECCV}}. \bibinfo{pages}{172--189}.
\newblock


\bibitem[Karayev et~al\mbox{.}(2014)]%
        {karayev2013recognizing}
\bibfield{author}{\bibinfo{person}{Sergey Karayev}, \bibinfo{person}{Matthew
  Trentacoste}, \bibinfo{person}{Helen Han}, \bibinfo{person}{Aseem Agarwala},
  \bibinfo{person}{Trevor Darrell}, \bibinfo{person}{Aaron Hertzmann}, {and}
  \bibinfo{person}{Holger Winnemoeller}.} \bibinfo{year}{2014}\natexlab{}.
\newblock \showarticletitle{Recognizing image style}. In
  \bibinfo{booktitle}{\emph{BMVC}}.
\newblock


\bibitem[Kazemi et~al\mbox{.}(2019)]%
        {kazemi2019style}
\bibfield{author}{\bibinfo{person}{Hadi Kazemi}, \bibinfo{person}{Seyed~Mehdi
  Iranmanesh}, {and} \bibinfo{person}{Nasser Nasrabadi}.}
  \bibinfo{year}{2019}\natexlab{}.
\newblock \showarticletitle{Style and content disentanglement in generative
  adversarial networks}. In \bibinfo{booktitle}{\emph{WACV}}. IEEE,
  \bibinfo{pages}{848--856}.
\newblock


\bibitem[Khan and van Noord(2021)]%
        {khan2021stylistic}
\bibfield{author}{\bibinfo{person}{Selina~J. Khan} {and} \bibinfo{person}{Nanne
  van Noord}.} \bibinfo{year}{2021}\natexlab{}.
\newblock \showarticletitle{Stylistic Multi-Task Analysis of Ukiyo-e Woodblock
  Prints}. In \bibinfo{booktitle}{\emph{BMVC}}. \bibinfo{pages}{1--5}.
\newblock


\bibitem[Khrulkov et~al\mbox{.}(2022)]%
        {khrulkov2021disentangled}
\bibfield{author}{\bibinfo{person}{Valentin Khrulkov}, \bibinfo{person}{Leyla
  Mirvakhabova}, \bibinfo{person}{Ivan Oseledets}, {and} \bibinfo{person}{Artem
  Babenko}.} \bibinfo{year}{2022}\natexlab{}.
\newblock \showarticletitle{Disentangled representations from non-disentangled
  models}.
\newblock \bibinfo{journal}{\emph{ICLR}} (\bibinfo{year}{2022}).
\newblock


\bibitem[Kim et~al\mbox{.}(2022)]%
        {kim2022diffusionclip}
\bibfield{author}{\bibinfo{person}{Gwanghyun Kim}, \bibinfo{person}{Taesung
  Kwon}, {and} \bibinfo{person}{Jong~Chul Ye}.}
  \bibinfo{year}{2022}\natexlab{}.
\newblock \showarticletitle{Diffusion{CLIP}: Text-Guided Diffusion Models for
  Robust Image Manipulation}. In \bibinfo{booktitle}{\emph{CVPR}}.
  \bibinfo{pages}{2426--2435}.
\newblock


\bibitem[Kingma and Ba(2015)]%
        {kingma2014adam}
\bibfield{author}{\bibinfo{person}{Diederick~P Kingma} {and}
  \bibinfo{person}{Jimmy Ba}.} \bibinfo{year}{2015}\natexlab{}.
\newblock \showarticletitle{Adam: A method for stochastic optimization}. In
  \bibinfo{booktitle}{\emph{ICLR}}.
\newblock


\bibitem[Kingma and Welling(2014)]%
        {kingma2013auto}
\bibfield{author}{\bibinfo{person}{Diederik~P Kingma} {and}
  \bibinfo{person}{Max Welling}.} \bibinfo{year}{2014}\natexlab{}.
\newblock \showarticletitle{Auto-encoding variational bayes}. In
  \bibinfo{booktitle}{\emph{ICLR}}.
\newblock


\bibitem[Kotovenko et~al\mbox{.}(2019)]%
        {kotovenko2019content}
\bibfield{author}{\bibinfo{person}{Dmytro Kotovenko}, \bibinfo{person}{Artsiom
  Sanakoyeu}, \bibinfo{person}{Sabine Lang}, {and} \bibinfo{person}{Bjorn
  Ommer}.} \bibinfo{year}{2019}\natexlab{}.
\newblock \showarticletitle{Content and style disentanglement for artistic
  style transfer}. In \bibinfo{booktitle}{\emph{ICCV}}.
  \bibinfo{pages}{4422--4431}.
\newblock


\bibitem[Kwon and Ye(2022)]%
        {kwon2022clipstyler}
\bibfield{author}{\bibinfo{person}{Gihyun Kwon} {and}
  \bibinfo{person}{Jong~Chul Ye}.} \bibinfo{year}{2022}\natexlab{}.
\newblock \showarticletitle{{CLIP}styler: Image style transfer with a single
  text condition}. In \bibinfo{booktitle}{\emph{CVPR}}.
  \bibinfo{pages}{18062--18071}.
\newblock


\bibitem[Kwon and Ye(2023)]%
        {kwon2022diffusion}
\bibfield{author}{\bibinfo{person}{Gihyun Kwon} {and}
  \bibinfo{person}{Jong~Chul Ye}.} \bibinfo{year}{2023}\natexlab{}.
\newblock \showarticletitle{Diffusion-based image translation using
  disentangled style and content representation}.
\newblock \bibinfo{journal}{\emph{ICLR}} (\bibinfo{year}{2023}).
\newblock


\bibitem[Lang and Ommer(2018)]%
        {lang2018reflecting}
\bibfield{author}{\bibinfo{person}{Sabine Lang} {and} \bibinfo{person}{Bjorn
  Ommer}.} \bibinfo{year}{2018}\natexlab{}.
\newblock \showarticletitle{Reflecting on how artworks are processed and
  analyzed by computer vision}. In \bibinfo{booktitle}{\emph{ECCV Workshops}}.
  \bibinfo{pages}{0--0}.
\newblock


\bibitem[Lecoutre et~al\mbox{.}(2017)]%
        {lecoutre2017recognizing}
\bibfield{author}{\bibinfo{person}{Adrian Lecoutre}, \bibinfo{person}{Benjamin
  Negrevergne}, {and} \bibinfo{person}{Florian Yger}.}
  \bibinfo{year}{2017}\natexlab{}.
\newblock \showarticletitle{Recognizing art style automatically in painting
  with deep learning}. In \bibinfo{booktitle}{\emph{Asian Conference on Machine
  Learning}}. PMLR, \bibinfo{pages}{327--342}.
\newblock


\bibitem[Li et~al\mbox{.}(2022)]%
        {li2022stylet2i}
\bibfield{author}{\bibinfo{person}{Zhiheng Li},
  \bibinfo{person}{Martin~Renqiang Min}, \bibinfo{person}{Kai Li}, {and}
  \bibinfo{person}{Chenliang Xu}.} \bibinfo{year}{2022}\natexlab{}.
\newblock \showarticletitle{Style{T}2{I}: Toward Compositional and
  High-Fidelity Text-to-Image Synthesis}. In \bibinfo{booktitle}{\emph{CVPR}}.
  \bibinfo{pages}{18197--18207}.
\newblock


\bibitem[Liao et~al\mbox{.}(2022)]%
        {liao2022text}
\bibfield{author}{\bibinfo{person}{Wentong Liao}, \bibinfo{person}{Kai Hu},
  \bibinfo{person}{Michael~Ying Yang}, {and} \bibinfo{person}{Bodo Rosenhahn}.}
  \bibinfo{year}{2022}\natexlab{}.
\newblock \showarticletitle{Text to image generation with semantic-spatial
  aware {GAN}}. In \bibinfo{booktitle}{\emph{CVPR}}.
  \bibinfo{pages}{18187--18196}.
\newblock


\bibitem[Liu et~al\mbox{.}(2022)]%
        {liu2022pseudo}
\bibfield{author}{\bibinfo{person}{Luping Liu}, \bibinfo{person}{Yi Ren},
  \bibinfo{person}{Zhijie Lin}, {and} \bibinfo{person}{Zhou Zhao}.}
  \bibinfo{year}{2022}\natexlab{}.
\newblock \showarticletitle{Pseudo numerical methods for diffusion models on
  manifolds}.
\newblock \bibinfo{journal}{\emph{ICLR}} (\bibinfo{year}{2022}).
\newblock


\bibitem[Liu et~al\mbox{.}(2021)]%
        {liu2020measuring}
\bibfield{author}{\bibinfo{person}{Xiao Liu}, \bibinfo{person}{Spyridon
  Thermos}, \bibinfo{person}{Gabriele Valvano}, \bibinfo{person}{Agisilaos
  Chartsias}, \bibinfo{person}{Alison O'Neil}, {and}
  \bibinfo{person}{Sotirios~A Tsaftaris}.} \bibinfo{year}{2021}\natexlab{}.
\newblock \showarticletitle{Measuring the Biases and Effectiveness of
  Content-Style Disentanglement}.
\newblock \bibinfo{journal}{\emph{BMVC}} (\bibinfo{year}{2021}).
\newblock


\bibitem[Ma et~al\mbox{.}(2017)]%
        {ma2017part}
\bibfield{author}{\bibinfo{person}{Daiqian Ma}, \bibinfo{person}{Feng Gao},
  \bibinfo{person}{Yan Bai}, \bibinfo{person}{Yihang Lou},
  \bibinfo{person}{Shiqi Wang}, \bibinfo{person}{Tiejun Huang}, {and}
  \bibinfo{person}{Ling-Yu Duan}.} \bibinfo{year}{2017}\natexlab{}.
\newblock \showarticletitle{From part to whole: who is behind the painting?}.
  In \bibinfo{booktitle}{\emph{ACM MM}}. \bibinfo{pages}{1174--1182}.
\newblock


\bibitem[Mao et~al\mbox{.}(2017)]%
        {mao2017deepart}
\bibfield{author}{\bibinfo{person}{Hui Mao}, \bibinfo{person}{Ming Cheung},
  {and} \bibinfo{person}{James She}.} \bibinfo{year}{2017}\natexlab{}.
\newblock \showarticletitle{Deep{A}rt: Learning joint representations of visual
  arts}. In \bibinfo{booktitle}{\emph{ACM MM}}. \bibinfo{pages}{1183--1191}.
\newblock


\bibitem[Mensink and Van~Gemert(2014)]%
        {mensink2014rijksmuseum}
\bibfield{author}{\bibinfo{person}{Thomas Mensink} {and} \bibinfo{person}{Jan
  Van~Gemert}.} \bibinfo{year}{2014}\natexlab{}.
\newblock \showarticletitle{The rijksmuseum challenge: Museum-centered visual
  recognition}. In \bibinfo{booktitle}{\emph{ICMR}}. \bibinfo{pages}{451--454}.
\newblock


\bibitem[Radford et~al\mbox{.}(2021)]%
        {radford2021learning}
\bibfield{author}{\bibinfo{person}{Alec Radford}, \bibinfo{person}{Jong~Wook
  Kim}, \bibinfo{person}{Chris Hallacy}, \bibinfo{person}{Aditya Ramesh},
  \bibinfo{person}{Gabriel Goh}, \bibinfo{person}{Sandhini Agarwal},
  \bibinfo{person}{Girish Sastry}, \bibinfo{person}{Amanda Askell},
  \bibinfo{person}{Pamela Mishkin}, \bibinfo{person}{Jack Clark},
  {et~al\mbox{.}}} \bibinfo{year}{2021}\natexlab{}.
\newblock \showarticletitle{Learning transferable visual models from natural
  language supervision}. In \bibinfo{booktitle}{\emph{ICML}}. PMLR,
  \bibinfo{pages}{8748--8763}.
\newblock


\bibitem[Ramesh et~al\mbox{.}(2022)]%
        {ramesh2022hierarchical}
\bibfield{author}{\bibinfo{person}{Aditya Ramesh}, \bibinfo{person}{Prafulla
  Dhariwal}, \bibinfo{person}{Alex Nichol}, \bibinfo{person}{Casey Chu}, {and}
  \bibinfo{person}{Mark Chen}.} \bibinfo{year}{2022}\natexlab{}.
\newblock \showarticletitle{Hierarchical text-conditional image generation with
  {CLIP} latents}.
\newblock \bibinfo{journal}{\emph{arXiv preprint arXiv:2204.06125}}
  (\bibinfo{year}{2022}).
\newblock


\bibitem[Rombach et~al\mbox{.}(2022)]%
        {rombach2022high}
\bibfield{author}{\bibinfo{person}{Robin Rombach}, \bibinfo{person}{Andreas
  Blattmann}, \bibinfo{person}{Dominik Lorenz}, \bibinfo{person}{Patrick
  Esser}, {and} \bibinfo{person}{Bj{\"o}rn Ommer}.}
  \bibinfo{year}{2022}\natexlab{}.
\newblock \showarticletitle{High-resolution image synthesis with latent
  diffusion models}. In \bibinfo{booktitle}{\emph{CVPR}}.
  \bibinfo{pages}{10684--10695}.
\newblock


\bibitem[Ruta et~al\mbox{.}(2021)]%
        {ruta2021aladin}
\bibfield{author}{\bibinfo{person}{Dan Ruta}, \bibinfo{person}{Saeid Motiian},
  \bibinfo{person}{Baldo Faieta}, \bibinfo{person}{Zhe Lin},
  \bibinfo{person}{Hailin Jin}, \bibinfo{person}{Alex Filipkowski},
  \bibinfo{person}{Andrew Gilbert}, {and} \bibinfo{person}{John Collomosse}.}
  \bibinfo{year}{2021}\natexlab{}.
\newblock \showarticletitle{{ALADIN}: all layer adaptive instance normalization
  for fine-grained style similarity}. In \bibinfo{booktitle}{\emph{ICCV}}.
  \bibinfo{pages}{11926--11935}.
\newblock


\bibitem[Sabatelli et~al\mbox{.}(2018)]%
        {sabatelli2018deep}
\bibfield{author}{\bibinfo{person}{Matthia Sabatelli}, \bibinfo{person}{Mike
  Kestemont}, \bibinfo{person}{Walter Daelemans}, {and} \bibinfo{person}{Pierre
  Geurts}.} \bibinfo{year}{2018}\natexlab{}.
\newblock \showarticletitle{Deep transfer learning for art classification
  problems}. In \bibinfo{booktitle}{\emph{ECCV Workshops}}.
  \bibinfo{pages}{0--0}.
\newblock


\bibitem[Saharia et~al\mbox{.}(2022)]%
        {saharia2022photorealistic}
\bibfield{author}{\bibinfo{person}{Chitwan Saharia}, \bibinfo{person}{William
  Chan}, \bibinfo{person}{Saurabh Saxena}, \bibinfo{person}{Lala Li},
  \bibinfo{person}{Jay Whang}, \bibinfo{person}{Emily~L Denton},
  \bibinfo{person}{Kamyar Ghasemipour}, \bibinfo{person}{Raphael
  Gontijo~Lopes}, \bibinfo{person}{Burcu Karagol~Ayan}, \bibinfo{person}{Tim
  Salimans}, {et~al\mbox{.}}} \bibinfo{year}{2022}\natexlab{}.
\newblock \showarticletitle{Photorealistic text-to-image diffusion models with
  deep language understanding}.
\newblock \bibinfo{journal}{\emph{NeurIPS}}  \bibinfo{volume}{35}
  (\bibinfo{year}{2022}), \bibinfo{pages}{36479--36494}.
\newblock


\bibitem[Saleh and Elgammal(2016)]%
        {saleh2016large}
\bibfield{author}{\bibinfo{person}{Babak Saleh} {and} \bibinfo{person}{Ahmed
  Elgammal}.} \bibinfo{year}{2016}\natexlab{}.
\newblock \showarticletitle{Large-scale Classification of Fine-Art Paintings:
  Learning The Right Metric on The Right Feature}.
\newblock \bibinfo{journal}{\emph{International Journal for Digital Art
  History}} \bibinfo{number}{2} (\bibinfo{year}{2016}).
\newblock


\bibitem[Sandoval et~al\mbox{.}(2019)]%
        {sandoval2019two}
\bibfield{author}{\bibinfo{person}{Catherine Sandoval}, \bibinfo{person}{Elena
  Pirogova}, {and} \bibinfo{person}{Margaret Lech}.}
  \bibinfo{year}{2019}\natexlab{}.
\newblock \showarticletitle{Two-stage deep learning approach to the
  classification of fine-art paintings}.
\newblock \bibinfo{journal}{\emph{IEEE Access}}  \bibinfo{volume}{7}
  (\bibinfo{year}{2019}), \bibinfo{pages}{41770--41781}.
\newblock


\bibitem[Sariyildiz et~al\mbox{.}(2023)]%
        {sariyildiz2023fake}
\bibfield{author}{\bibinfo{person}{Mert~Bulent Sariyildiz},
  \bibinfo{person}{Karteek Alahari}, \bibinfo{person}{Diane Larlus}, {and}
  \bibinfo{person}{Yannis Kalantidis}.} \bibinfo{year}{2023}\natexlab{}.
\newblock \showarticletitle{Fake it till you make it: Learning transferable
  representations from synthetic ImageNet clones}. In
  \bibinfo{booktitle}{\emph{CVPR}}.
\newblock


\bibitem[Schroff et~al\mbox{.}(2015)]%
        {schroff2015facenet}
\bibfield{author}{\bibinfo{person}{Florian Schroff}, \bibinfo{person}{Dmitry
  Kalenichenko}, {and} \bibinfo{person}{James Philbin}.}
  \bibinfo{year}{2015}\natexlab{}.
\newblock \showarticletitle{Face{N}et: A unified embedding for face recognition
  and clustering}. In \bibinfo{booktitle}{\emph{CVPR}}.
  \bibinfo{pages}{815--823}.
\newblock


\bibitem[Schuhmann et~al\mbox{.}(2022)]%
        {schuhmann2022laion}
\bibfield{author}{\bibinfo{person}{Christoph Schuhmann},
  \bibinfo{person}{Romain Beaumont}, \bibinfo{person}{Richard Vencu},
  \bibinfo{person}{Cade Gordon}, \bibinfo{person}{Ross Wightman},
  \bibinfo{person}{Mehdi Cherti}, \bibinfo{person}{Theo Coombes},
  \bibinfo{person}{Aarush Katta}, \bibinfo{person}{Clayton Mullis},
  \bibinfo{person}{Mitchell Wortsman}, {et~al\mbox{.}}}
  \bibinfo{year}{2022}\natexlab{}.
\newblock \showarticletitle{L{AION-5B}: An open large-scale dataset for
  training next generation image-text models}.
\newblock \bibinfo{journal}{\emph{NeurIPS}} (\bibinfo{year}{2022}).
\newblock


\bibitem[Shao et~al\mbox{.}(2020)]%
        {shao2020controlvae}
\bibfield{author}{\bibinfo{person}{Huajie Shao}, \bibinfo{person}{Shuochao
  Yao}, \bibinfo{person}{Dachun Sun}, \bibinfo{person}{Aston Zhang},
  \bibinfo{person}{Shengzhong Liu}, \bibinfo{person}{Dongxin Liu},
  \bibinfo{person}{Jun Wang}, {and} \bibinfo{person}{Tarek Abdelzaher}.}
  \bibinfo{year}{2020}\natexlab{}.
\newblock \showarticletitle{Control{VAE}: Controllable variational
  autoencoder}. In \bibinfo{booktitle}{\emph{ICML}}. PMLR,
  \bibinfo{pages}{8655--8664}.
\newblock


\bibitem[Shen et~al\mbox{.}(2019)]%
        {shen2019discovering}
\bibfield{author}{\bibinfo{person}{Xi Shen}, \bibinfo{person}{Alexei~A Efros},
  {and} \bibinfo{person}{Mathieu Aubry}.} \bibinfo{year}{2019}\natexlab{}.
\newblock \showarticletitle{Discovering visual patterns in art collections with
  spatially-consistent feature learning}. In \bibinfo{booktitle}{\emph{CVPR}}.
  \bibinfo{pages}{9278--9287}.
\newblock


\bibitem[Shi et~al\mbox{.}(2022)]%
        {shi2022semanticstylegan}
\bibfield{author}{\bibinfo{person}{Yichun Shi}, \bibinfo{person}{Xiao Yang},
  \bibinfo{person}{Yangyue Wan}, {and} \bibinfo{person}{Xiaohui Shen}.}
  \bibinfo{year}{2022}\natexlab{}.
\newblock \showarticletitle{Semantic{S}tyle{GAN}: Learning Compositional
  Generative Priors for Controllable Image Synthesis and Editing}. In
  \bibinfo{booktitle}{\emph{CVPR}}. \bibinfo{pages}{11254--11264}.
\newblock


\bibitem[Simonyan and Zisserman(2015)]%
        {simonyan2014very}
\bibfield{author}{\bibinfo{person}{Karen Simonyan} {and}
  \bibinfo{person}{Andrew Zisserman}.} \bibinfo{year}{2015}\natexlab{}.
\newblock \showarticletitle{Very deep convolutional networks for large-scale
  image recognition}. In \bibinfo{booktitle}{\emph{ICLR}}.
\newblock


\bibitem[Sohn(2016)]%
        {sohn2016improved}
\bibfield{author}{\bibinfo{person}{Kihyuk Sohn}.}
  \bibinfo{year}{2016}\natexlab{}.
\newblock \showarticletitle{Improved deep metric learning with multi-class
  n-pair loss objective}.
\newblock \bibinfo{journal}{\emph{Advances in neural information processing
  systems}}  \bibinfo{volume}{29} (\bibinfo{year}{2016}).
\newblock


\bibitem[Strezoski and Worring(2018)]%
        {strezoski2018omniart}
\bibfield{author}{\bibinfo{person}{Gjorgji Strezoski} {and}
  \bibinfo{person}{Marcel Worring}.} \bibinfo{year}{2018}\natexlab{}.
\newblock \showarticletitle{Omni{A}rt: a large-scale artistic benchmark}.
\newblock \bibinfo{journal}{\emph{TOMM}} \bibinfo{volume}{14},
  \bibinfo{number}{4} (\bibinfo{year}{2018}), \bibinfo{pages}{1--21}.
\newblock


\bibitem[Tan et~al\mbox{.}(2020)]%
        {tan2020kt}
\bibfield{author}{\bibinfo{person}{Hongchen Tan}, \bibinfo{person}{Xiuping
  Liu}, \bibinfo{person}{Meng Liu}, \bibinfo{person}{Baocai Yin}, {and}
  \bibinfo{person}{Xin Li}.} \bibinfo{year}{2020}\natexlab{}.
\newblock \showarticletitle{{KT-GAN}: knowledge-transfer generative adversarial
  network for text-to-image synthesis}.
\newblock \bibinfo{journal}{\emph{Transactions on Image Processing}}
  \bibinfo{volume}{30} (\bibinfo{year}{2020}), \bibinfo{pages}{1275--1290}.
\newblock


\bibitem[Tan et~al\mbox{.}(2019)]%
        {artgan2018}
\bibfield{author}{\bibinfo{person}{Wei~Ren Tan}, \bibinfo{person}{Chee~Seng
  Chan}, \bibinfo{person}{Hernan Aguirre}, {and} \bibinfo{person}{Kiyoshi
  Tanaka}.} \bibinfo{year}{2019}\natexlab{}.
\newblock \showarticletitle{Improved {A}rt{GAN} for Conditional Synthesis of
  Natural Image and Artwork}.
\newblock \bibinfo{journal}{\emph{Transactions on Image Processing}}
  \bibinfo{volume}{28}, \bibinfo{number}{1} (\bibinfo{year}{2019}),
  \bibinfo{pages}{394--409}.
\newblock
\urldef\tempurl%
\url{https://doi.org/10.1109/TIP.2018.2866698}
\showDOI{\tempurl}


\bibitem[Tan et~al\mbox{.}(2016)]%
        {tan2016ceci}
\bibfield{author}{\bibinfo{person}{Wei~Ren Tan}, \bibinfo{person}{Chee~Seng
  Chan}, \bibinfo{person}{Hern{\'a}n~E Aguirre}, {and} \bibinfo{person}{Kiyoshi
  Tanaka}.} \bibinfo{year}{2016}\natexlab{}.
\newblock \showarticletitle{Ceci n'est pas une pipe: A deep convolutional
  network for fine-art paintings classification}. In
  \bibinfo{booktitle}{\emph{ICIP}}. IEEE, \bibinfo{pages}{3703--3707}.
\newblock


\bibitem[Tao et~al\mbox{.}(2022)]%
        {tao2022df}
\bibfield{author}{\bibinfo{person}{Ming Tao}, \bibinfo{person}{Hao Tang},
  \bibinfo{person}{Fei Wu}, \bibinfo{person}{Xiao-Yuan Jing},
  \bibinfo{person}{Bing-Kun Bao}, {and} \bibinfo{person}{Changsheng Xu}.}
  \bibinfo{year}{2022}\natexlab{}.
\newblock \showarticletitle{{DF-GAN}: A Simple and Effective Baseline for
  Text-to-Image Synthesis}. In \bibinfo{booktitle}{\emph{CVPR}}.
  \bibinfo{pages}{16515--16525}.
\newblock


\bibitem[Tran et~al\mbox{.}(2017)]%
        {tran2017disentangled}
\bibfield{author}{\bibinfo{person}{Luan Tran}, \bibinfo{person}{Xi Yin}, {and}
  \bibinfo{person}{Xiaoming Liu}.} \bibinfo{year}{2017}\natexlab{}.
\newblock \showarticletitle{Disentangled representation learning {GAN} for
  pose-invariant face recognition}. In \bibinfo{booktitle}{\emph{CVPR}}.
  \bibinfo{pages}{1415--1424}.
\newblock


\bibitem[Tumanyan et~al\mbox{.}(2022)]%
        {tumanyan2022splicing}
\bibfield{author}{\bibinfo{person}{Narek Tumanyan}, \bibinfo{person}{Omer
  Bar-Tal}, \bibinfo{person}{Shai Bagon}, {and} \bibinfo{person}{Tali Dekel}.}
  \bibinfo{year}{2022}\natexlab{}.
\newblock \showarticletitle{Splicing {ViT} Features for Semantic Appearance
  Transfer}. In \bibinfo{booktitle}{\emph{CVPR}}.
  \bibinfo{pages}{10748--10757}.
\newblock


\bibitem[Van~Noord et~al\mbox{.}(2015)]%
        {van2015toward}
\bibfield{author}{\bibinfo{person}{Nanne Van~Noord}, \bibinfo{person}{Ella
  Hendriks}, {and} \bibinfo{person}{Eric Postma}.}
  \bibinfo{year}{2015}\natexlab{}.
\newblock \showarticletitle{Toward discovery of the artist's style: Learning to
  recognize artists by their artworks}.
\newblock \bibinfo{journal}{\emph{IEEE Signal Processing Magazine}}
  \bibinfo{volume}{32}, \bibinfo{number}{4} (\bibinfo{year}{2015}),
  \bibinfo{pages}{46--54}.
\newblock


\bibitem[Wilber et~al\mbox{.}(2017)]%
        {wilber2017bam}
\bibfield{author}{\bibinfo{person}{Michael~J Wilber}, \bibinfo{person}{Chen
  Fang}, \bibinfo{person}{Hailin Jin}, \bibinfo{person}{Aaron Hertzmann},
  \bibinfo{person}{John Collomosse}, {and} \bibinfo{person}{Serge Belongie}.}
  \bibinfo{year}{2017}\natexlab{}.
\newblock \showarticletitle{{BAM}! The behance artistic media dataset for
  recognition beyond photography}. In \bibinfo{booktitle}{\emph{ICCV}}.
  \bibinfo{pages}{1202--1211}.
\newblock


\bibitem[Wu et~al\mbox{.}(2022)]%
        {wu2022uncovering}
\bibfield{author}{\bibinfo{person}{Qiucheng Wu}, \bibinfo{person}{Yujian Liu},
  \bibinfo{person}{Handong Zhao}, \bibinfo{person}{Ajinkya Kale},
  \bibinfo{person}{Trung Bui}, \bibinfo{person}{Tong Yu}, \bibinfo{person}{Zhe
  Lin}, \bibinfo{person}{Yang Zhang}, {and} \bibinfo{person}{Shiyu Chang}.}
  \bibinfo{year}{2022}\natexlab{}.
\newblock \showarticletitle{Uncovering the Disentanglement Capability in
  Text-to-Image Diffusion Models}.
\newblock \bibinfo{journal}{\emph{arXiv preprint arXiv:2212.08698}}
  (\bibinfo{year}{2022}).
\newblock


\bibitem[Xie et~al\mbox{.}(2022)]%
        {xie2022artistic}
\bibfield{author}{\bibinfo{person}{Xin Xie}, \bibinfo{person}{Yi Li},
  \bibinfo{person}{Huaibo Huang}, \bibinfo{person}{Haiyan Fu},
  \bibinfo{person}{Wanwan Wang}, {and} \bibinfo{person}{Yanqing Guo}.}
  \bibinfo{year}{2022}\natexlab{}.
\newblock \showarticletitle{Artistic Style Discovery With Independent
  Components}. In \bibinfo{booktitle}{\emph{CVPR}}.
  \bibinfo{pages}{19870--19879}.
\newblock


\bibitem[Xu et~al\mbox{.}(2022)]%
        {xu2022predict}
\bibfield{author}{\bibinfo{person}{Zipeng Xu}, \bibinfo{person}{Tianwei Lin},
  \bibinfo{person}{Hao Tang}, \bibinfo{person}{Fu Li},
  \bibinfo{person}{Dongliang He}, \bibinfo{person}{Nicu Sebe},
  \bibinfo{person}{Radu Timofte}, \bibinfo{person}{Luc Van~Gool}, {and}
  \bibinfo{person}{Errui Ding}.} \bibinfo{year}{2022}\natexlab{}.
\newblock \showarticletitle{Predict, Prevent, and Evaluate: Disentangled
  Text-Driven Image Manipulation Empowered by Pre-Trained Vision-Language
  Model}. In \bibinfo{booktitle}{\emph{CVPR}}. \bibinfo{pages}{18229--18238}.
\newblock


\bibitem[You et~al\mbox{.}(2020)]%
        {you2017large}
\bibfield{author}{\bibinfo{person}{Yang You}, \bibinfo{person}{Igor Gitman},
  {and} \bibinfo{person}{Boris Ginsburg}.} \bibinfo{year}{2020}\natexlab{}.
\newblock \showarticletitle{Large batch training of convolutional networks}.
\newblock \bibinfo{journal}{\emph{ICLR}} (\bibinfo{year}{2020}).
\newblock


\bibitem[Ypsilantis et~al\mbox{.}(2021)]%
        {ypsilantis2021met}
\bibfield{author}{\bibinfo{person}{Nikolaos-Antonios Ypsilantis},
  \bibinfo{person}{Noa Garcia}, \bibinfo{person}{Guangxing Han},
  \bibinfo{person}{Sarah Ibrahimi}, \bibinfo{person}{Nanne Van~Noord}, {and}
  \bibinfo{person}{Giorgos Tolias}.} \bibinfo{year}{2021}\natexlab{}.
\newblock \showarticletitle{The {M}et dataset: Instance-level recognition for
  artworks}. In \bibinfo{booktitle}{\emph{NeurIPS Datasets and Benchmarks
  Track}}.
\newblock


\bibitem[Yu et~al\mbox{.}(2019)]%
        {yu2019multi}
\bibfield{author}{\bibinfo{person}{Xiaoming Yu}, \bibinfo{person}{Yuanqi Chen},
  \bibinfo{person}{Shan Liu}, \bibinfo{person}{Thomas Li}, {and}
  \bibinfo{person}{Ge Li}.} \bibinfo{year}{2019}\natexlab{}.
\newblock \showarticletitle{Multi-mapping image-to-image translation via
  learning disentanglement}.
\newblock \bibinfo{journal}{\emph{NeurIPS}}  \bibinfo{volume}{32}
  (\bibinfo{year}{2019}).
\newblock


\bibitem[Zhang et~al\mbox{.}(2022)]%
        {zhang2022pointclip}
\bibfield{author}{\bibinfo{person}{Renrui Zhang}, \bibinfo{person}{Ziyu Guo},
  \bibinfo{person}{Wei Zhang}, \bibinfo{person}{Kunchang Li},
  \bibinfo{person}{Xupeng Miao}, \bibinfo{person}{Bin Cui}, \bibinfo{person}{Yu
  Qiao}, \bibinfo{person}{Peng Gao}, {and} \bibinfo{person}{Hongsheng Li}.}
  \bibinfo{year}{2022}\natexlab{}.
\newblock \showarticletitle{Point{CLIP}: Point cloud understanding by {CLIP}}.
  In \bibinfo{booktitle}{\emph{CVPR}}. \bibinfo{pages}{8552--8562}.
\newblock


\bibitem[Zhou et~al\mbox{.}(2022)]%
        {zhou2022towards}
\bibfield{author}{\bibinfo{person}{Yufan Zhou}, \bibinfo{person}{Ruiyi Zhang},
  \bibinfo{person}{Changyou Chen}, \bibinfo{person}{Chunyuan Li},
  \bibinfo{person}{Chris Tensmeyer}, \bibinfo{person}{Tong Yu},
  \bibinfo{person}{Jiuxiang Gu}, \bibinfo{person}{Jinhui Xu}, {and}
  \bibinfo{person}{Tong Sun}.} \bibinfo{year}{2022}\natexlab{}.
\newblock \showarticletitle{Towards Language-Free Training for Text-to-Image
  Generation}. In \bibinfo{booktitle}{\emph{CVPR}}.
  \bibinfo{pages}{17907--17917}.
\newblock


\end{thebibliography}

\newpage
\appendix

%%%%%%%%% BODY TEXT - ENTER YOUR RESPONSE BELOW
\section{GOYA details} \label{sec:sup:GOYAdetails}
The details of GOYA architecture are shown in Table \ref{tab:supple_GOYA_details}.
\begin{table}[h]
\renewcommand{\arraystretch}{1.1}
\small
\centering
\caption{GOYA detailed architecture.}
% \vspace{-5pt}
\begin{tabularx}{0.68\columnwidth}{@{}l l l}
\toprule
\multicolumn{1}{l}{Components} & Layer details \\
\midrule
\multicolumn{1}{l}{\multirow{3}{*}{Content encoder $\mathcal{C}$}} & Linear layer $(512, 2048)$ \\\cmidrule{2-2}
                                                                    & ReLU non-linearity \\\cmidrule{2-2}
                                                                    & Linear layer $(2048, 2048)$ \\\cmidrule{1-2}
\multicolumn{1}{l}{\multirow{5}{*}{Style encoder $\mathcal{S}$}} & Linear layer $(512, 512)$ \\\cmidrule{2-2}
                                                                    & ReLU non-linearity \\\cmidrule{2-2}
                                                                    & Linear layer $(512, 512)$ \\\cmidrule{2-2}
                                                                    & ReLU non-linearity \\\cmidrule{2-2}
                                                                    & Linear layer $(512, 2048)$ \\\cmidrule{1-2}
\multicolumn{1}{l}{\multirow{3}{*}{Projector $h^C/h^S$}} & Linear layer $(2048, 2048)$ \\\cmidrule{2-2}
                                                            & ReLU non-linearity \\\cmidrule{2-2}
                                                            & Linear layer $(2048, 64)$ \\\cmidrule{1-2}
\multicolumn{1}{l}{Style classifier $\mathcal{R}$} & Linear layer $(2048, 27)$ \\

\bottomrule
\end{tabularx}
\label{tab:supple_GOYA_details}
\vspace{-10pt}
\end{table}

\section{Baseline details} \label{sec:sup:bslinedetails}

\paragraph{Fine-tuning ResNet50 and CLIP}
ResNet50 and CLIP are fine-tuned by adding a linear classifier for genre or style movement after the layer from which embeddings are extracted, and then training the entire model on top of the pre-trained checkpoint. The ground-truth is the genre or style movement label in the WikiArt, or style movement in the prompt of diffusion-generated images.

\paragraph{Embeddings of baselines}
Gram matrix embeddings are computed from the layer \textit{conv5\_1} of a pre-trained VGG19 \cite{simonyan2014very}. For ResNet50 \cite{He2016DeepRL}, CLIP \cite{radford2021learning} and DINO \cite{caron2021emerging}, the protocols for which layer to extract embeddings and for fine-tuning are consistent as in the disentanglement task.

\section{Classification evaluation details} \label{sec:sup:clsevadetails}

\paragraph{Labels in classification evaluation}
We use $10$ genres and $27$ style movements in the WikiArt \cite{artgan2018} dataset for classification evaluation. 

Genre labels include: \textit{abstract painting, cityscape, genre painting, illustration, landscape, nude painting, portrait, sketch and study, religious painting} and \textit{still life}. 

Style movement labels include: \textit{Abstract Expressionism, Action painting, Analytical Cubism, Art Nouveau, Baroque, Color Field Painting, Contemporary Realism, Cubism, Early Renaissance, Expressionism, Fauvism, High Renaissance, Impressionism, Mannerism Late Renaissance, Minimalism, Naive Art Primitivism, New Realism, Northern Renaissance, Pointillism, Pop Art, Post Impressionism, Realism, Rococo, Romanticism, Symbolism, Synthetic Cubism} and \textit{Ukiyo-e.}

\paragraph{Classifier training details}
The optimizer is LARS \cite{you2017large} with initial learning rate $0.02$, a cosine delay schedule, and momentum $=0.9$. The batch size is $4,096$. We train each classifier for $90$ epochs.

\begin{figure*}
\hspace{-25pt}
    \centering
    \includegraphics[width=0.65\textwidth]{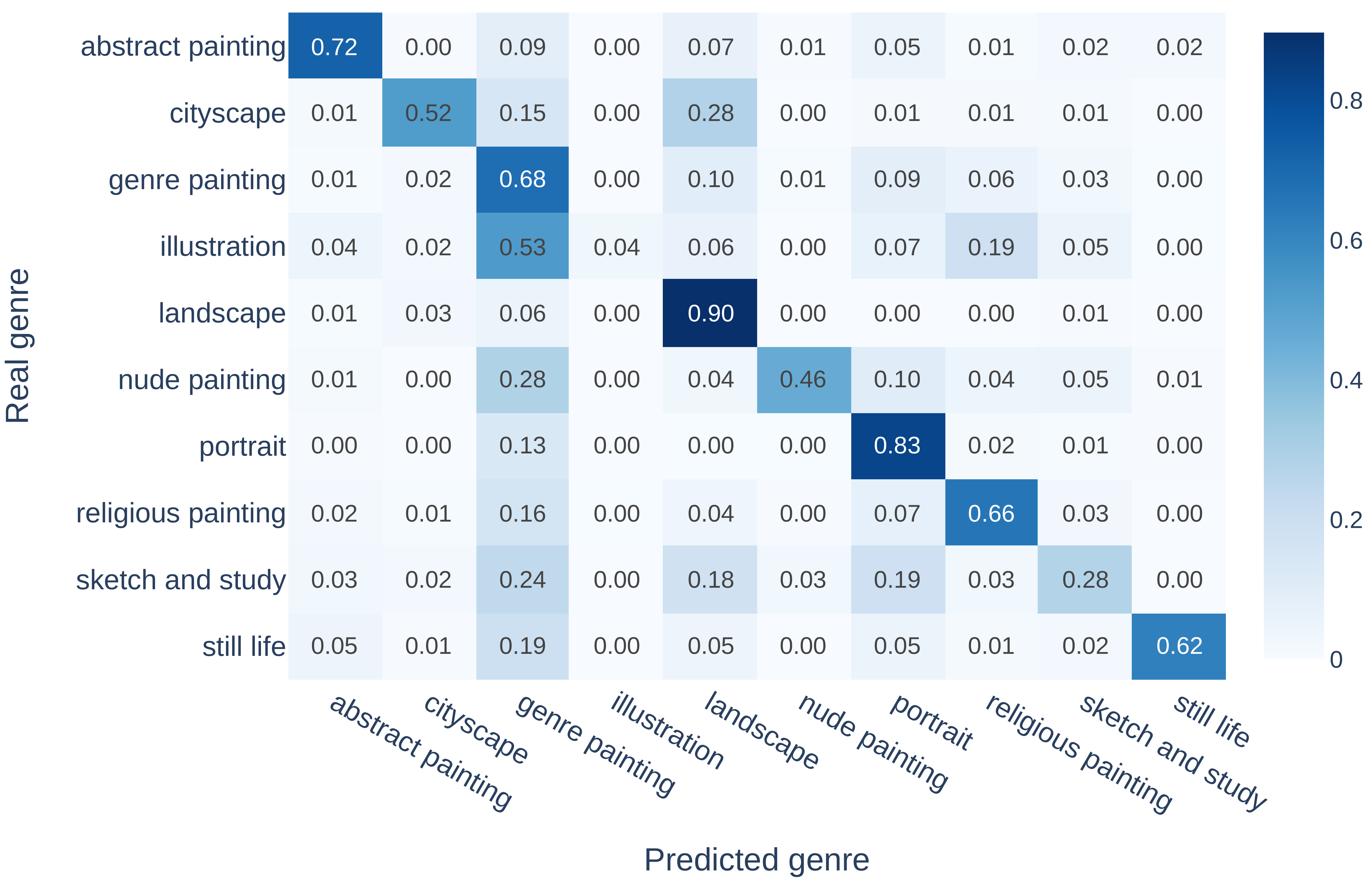}
    \caption{Confusion matrix of genre classification evaluation on GOYA content space.}
    \label{fig:supple_genre}
\end{figure*}

\paragraph{Confusion matrix}
Figure \ref{fig:supple_genre} shows the confusion matrix of genre classification evaluation on GOYA content space. The number in each cell represents the proportion of images that are classified as the predicted label to the total images with the true label. The darker the color, the more images are classified as the predicted label. We can observe that images from several genres are misclassified as \textit{genre painting}, as \textit{genre painting}s usually depict a wide aboard of activities in daily life, thus have overlapping semantics with images from other genres, such as \textit{illustration} and \textit{nude painting}. In addition, due to the high similarity of the depicted scenes, there are $28\%$ of the images from \textit{cityscape} misclassified as \textit{landscape}. 

\begin{figure*}
\hspace{-25pt}
    \centering
    \includegraphics[width=1\textwidth]{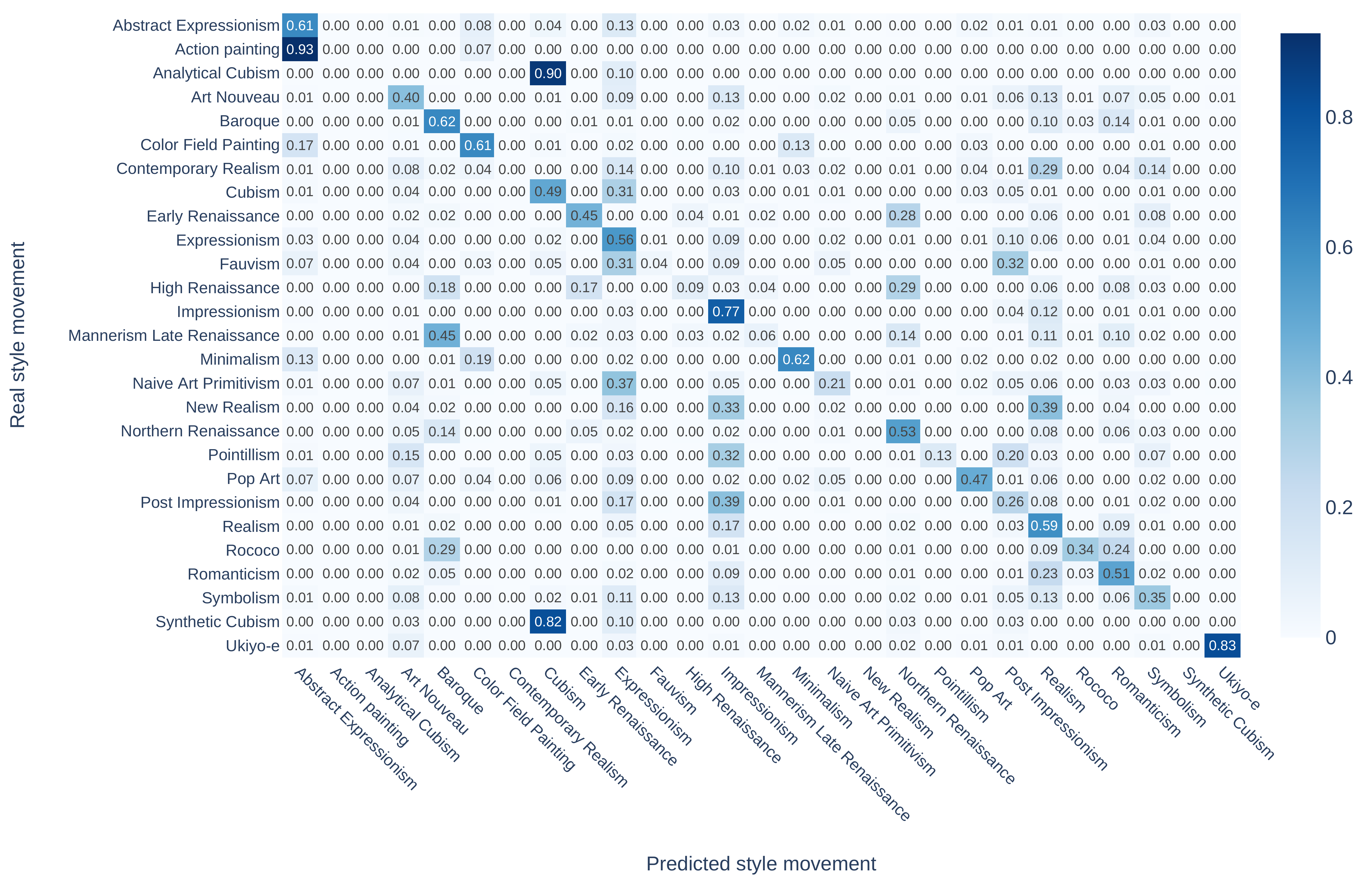}
    \caption{Confusion matrix of style movement classification evaluation on GOYA style space.}
    \label{fig:supple_style}
\end{figure*}

Figure \ref{fig:supple_style} shows the confusion matrix of style movement classification in GOYA style space. However, the boundary of some movements is not very clear, as some movements are sub-movements that represent different phases in one major movement, e.g. \textit{Synthetic Cubism} in \textit{Cubism} and \textit{Post Impressionism} in \textit{Impressionism}. Generative models may produce images likely to the major movement even if when the prompt is about sub-movements, leading GOYA to learn from inaccurate information. Thus, images from sub-movements are prone to be predicted as the according major movement. For example, $82\%$ of the images in \textit{Synthetic Cubism} and $90\%$ of the images in \textit{Analytical Cubism} are classified as \textit{Cubism}. Similarly, about $1/3$ of the images in \textit{Contemporary Realism} and \textit{New Realism} are predicted incorrectly as \textit{Realism}.

\section{Similarity retrieval comparison} \label{sec:sup:similarity}
\begin{figure*}
\hspace{-25pt}
    \centering
    \includegraphics[width=0.88\textwidth]{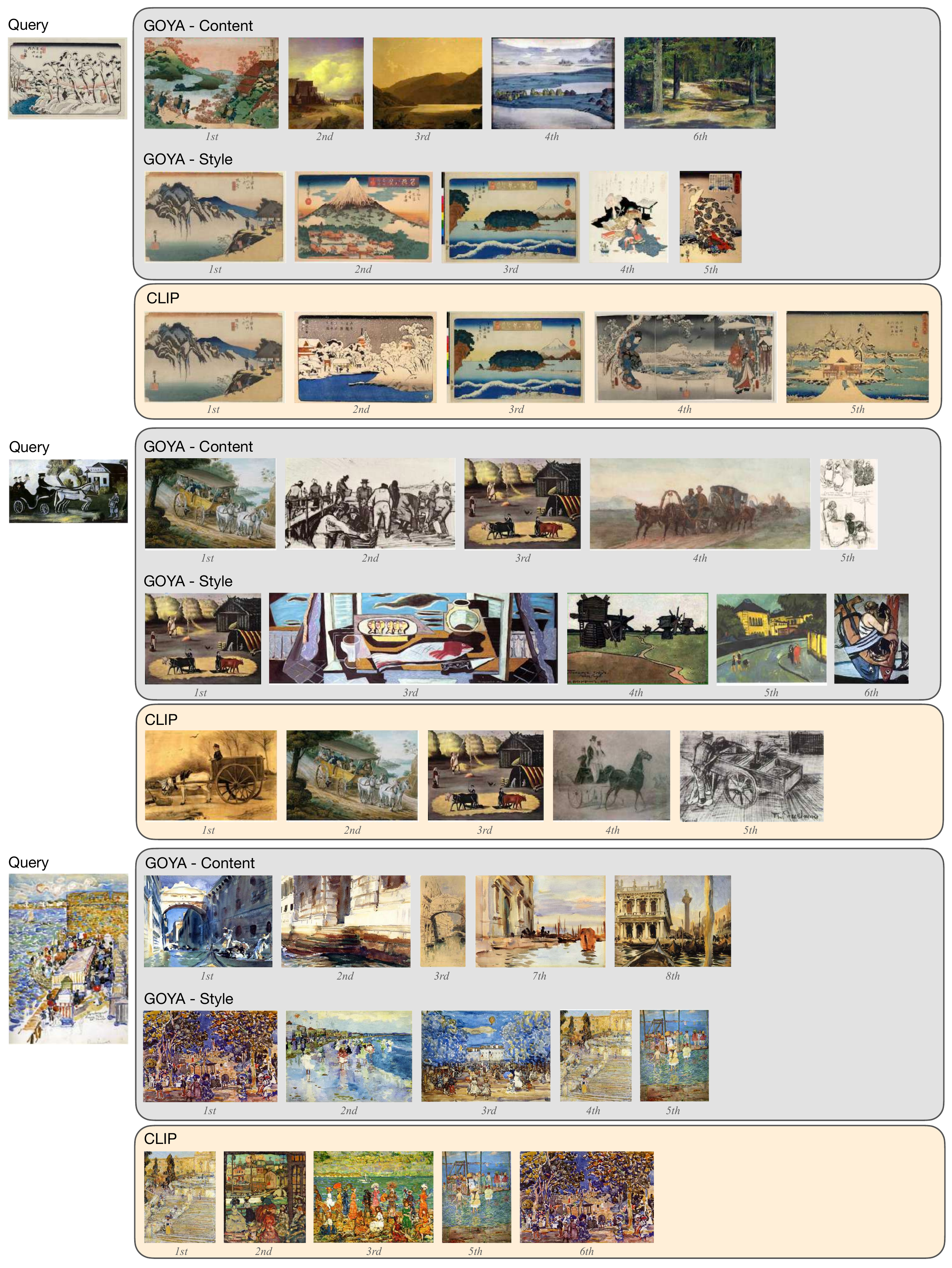}
    \caption{Retrieval results in GOYA content and style spaces and CLIP latent space based on cosine similarity. In each row, the similarity decreases from left to right. Copyrighted images are skipped.}
    \label{fig:supple_0}
\end{figure*}

\begin{figure*}
\hspace{-25pt}
    \centering
    \includegraphics[width=0.80\textwidth]{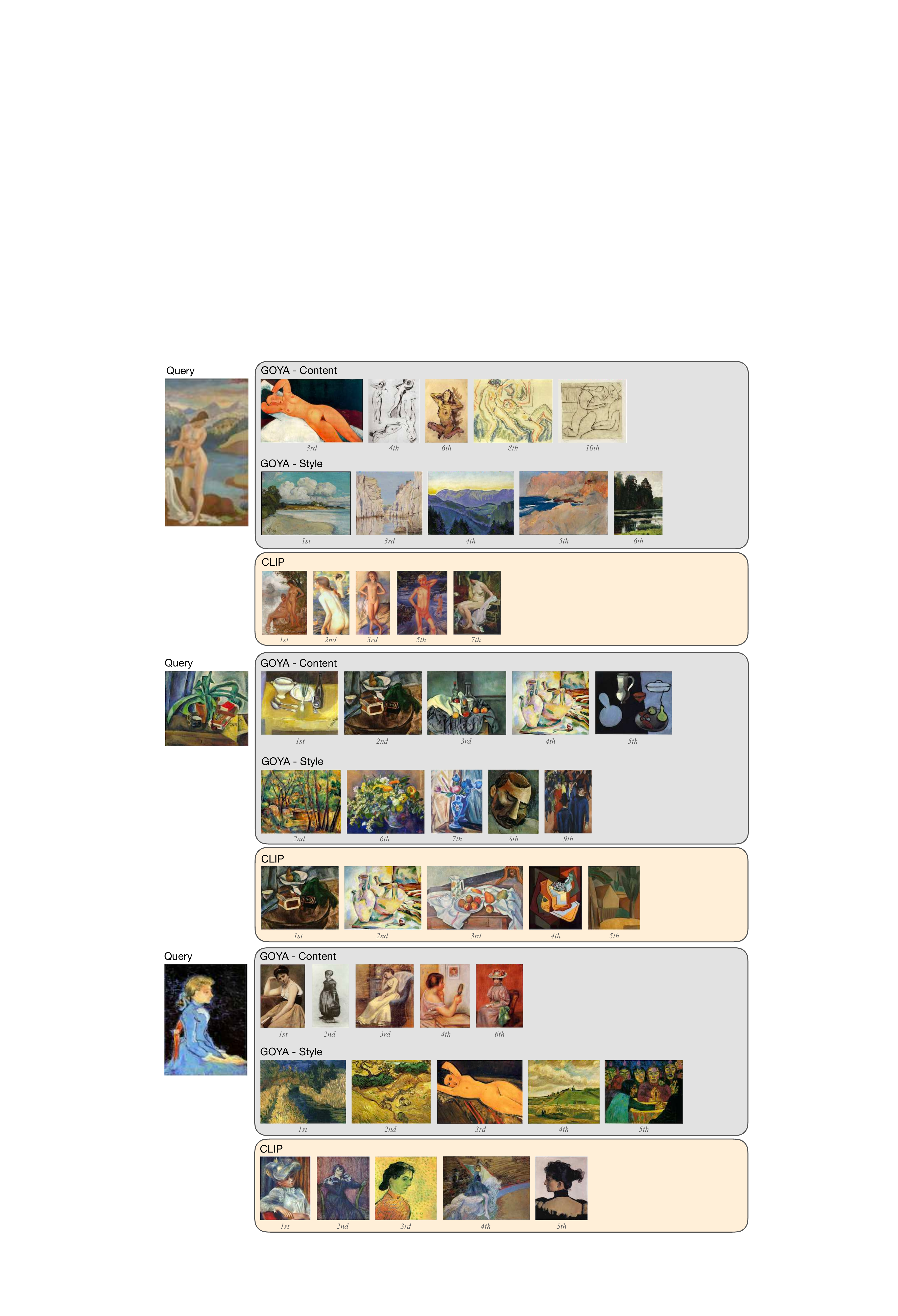}
    \caption{Retrieval results in GOYA content and style spaces and CLIP latent space based on cosine similarity. In each row, the similarity decreases from left to right. Copyrighted images are skipped.}
    \label{fig:supple_1}
\end{figure*}

Here we show more results (Figure \ref{fig:supple_0}, \ref{fig:supple_1}) on similarity retrieval and compare them against CLIP \cite{radford2021learning}. 
In Figure \ref{fig:supple_0} and \ref{fig:supple_1}, for each query image, the first two rows display the retrieved images from GOYA content and style spaces, and the last row shows images retrieved in the CLIP latent space. Results show that images in the CLIP latent space are similar in content and style, while in GOYA content space display consistency in depicting scenes but with different styles, and in GOYA style space the visual appearance is similar but the content is different.

\end{document}